\documentclass{article}

\usepackage{microtype}
\usepackage{graphicx}
\usepackage{subfigure}
\usepackage{booktabs} 

\usepackage{hyperref}

\usepackage[accepted]{icml2024}

\usepackage{amsmath}
\usepackage{amssymb}
\usepackage{mathtools}
\usepackage{amsthm}

\usepackage{bm}
\usepackage{threeparttable}
\usepackage{multirow}
\usepackage{wrapfig}
\usepackage{float}
\usepackage{makecell}

\usepackage[capitalize,noabbrev]{cleveref}

\theoremstyle{plain}

\theoremstyle{definition}

\theoremstyle{remark}

\usepackage[textsize=tiny]{todonotes}

\begin{document}

\twocolumn[
\icmltitle{Omni-Dimensional Frequency Learner for General Time Series Analysis}




\begin{icmlauthorlist}
\icmlauthor{Xianing Chen}{yyy}
\icmlauthor{Hanting Chen}{yyy}
\icmlauthor{Hailin Hu}{yyy}
\end{icmlauthorlist}

\icmlaffiliation{yyy}{Huawei Noah’s Ark Lab}


\vskip 0.3in
]



\printAffiliationsAndNotice{}  

\begin{abstract}
Frequency domain representation of time series feature offers a concise representation for handling real-world time series data with inherent complexity and dynamic nature. 
However, current frequency-based methods with complex operations still fall short of state-of-the-art time domain methods for general time series analysis. 
In this work, we present Omni-Dimensional Frequency Learner (ODFL) model based on a in depth analysis among all the three aspects of the spectrum feature: channel redundancy property among the frequency dimension, the sparse and un-salient frequency energy distribution among the frequency dimension, and the semantic diversity among the variable dimension. 
Technically, our method is composed of a semantic-adaptive global filter with attention to the un-salient frequency bands and partial operation among the channel dimension.
Empirical results show that ODFL achieves consistent state-of-the-art in five mainstream time series analysis tasks, including short- and long-term forecasting, imputation, classification, and anomaly detection, offering a promising foundation for time series analysis.
\end{abstract}

\section{Introduction}
\label{sec:intro}
Time series analysis, including the tasks of long/short-term forecasting \cite{Lim2021TimeseriesFW,M4team2018dataset}, imputation \cite{Friedman1962TheIO},
anomaly detection \cite{xu2021anomaly}, and classification \cite{Bagnall2018TheUM}, has immense practical value in real-world applications such as energy, finance, signal processing \cite{Taylor2018Forecasting}. Thus, it is crucial to design a general-purpose model for different tasks \cite{wu2023timesnet} other than focus on specific task in previous works \cite{woo2022etsformer}. However, it is challenging to model time series with inherent complexity and dynamic nature, especially for its sparse, mixing, and overlapped multiple variations \cite{zhou2022fedformer,wu2023timesnet}.

Fortunately, the frequency domain representation of time series data offers a more concise representation of its intricate information. It is much easier to model the intrinsic properties of time-series data, e.g. cyclic, trend, global interaction. Thus, FEDFormer \cite{zhou2022fedformer} enhances feature extracted from transformer-based model by incorporating spectral information. TimesNet \cite{wu2023timesnet} extract multi-periodicity feature by segmenting data based on frequencies. However, their proposed complex operations which focus on designing extra sophisticated structures to work with the traditional architecture makes them heavy-weight. Moreover, they still cannot achieve better performance to the state-of-the-art time domain methods \cite{nie2022time}.

This raise the questions of constructing frequency domain time series model: \emph{how to extract frequency domain feature gracefully? And how to extract the most informative part among the frequency domain feature?}

\begin{figure*}[ht]
	\begin{center}
		\subfigure[baseline]{
			\label{fig_1a}
			\centering
			\includegraphics[width=0.245\linewidth]{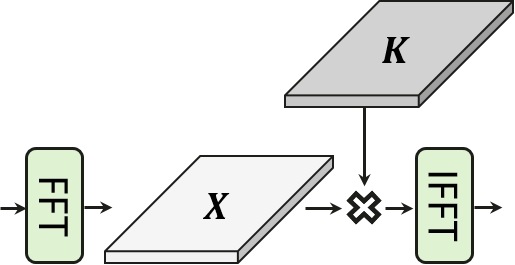}
		}
		\subfigure[channel dimension]{
			\label{fig_1b}
			\centering
			\includegraphics[width=0.25\linewidth]{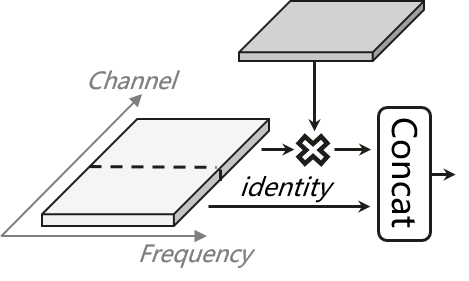}
		}
		\subfigure[frequency dimension]{
			\label{fig_1c}
			\centering
			\includegraphics[width=0.185\linewidth]{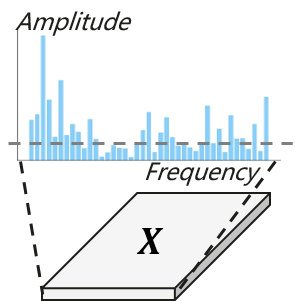}
		}
		\subfigure[variable dimension]{
			\label{fig_1d}
			\centering
			\includegraphics[width=0.24\linewidth]{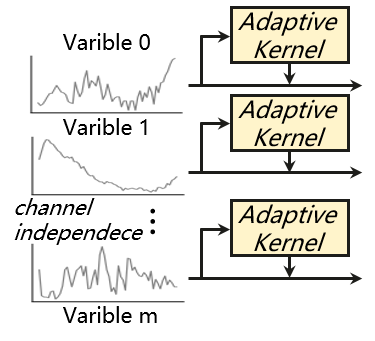}
		}
	\end{center}
	\caption{Illustration of (a) the baseline operator, (b) the partial operation, (c) the unsalient frequency bands feature extraction, (d) the semantic diversity and adaptation under the channel independent setting.}
	\label{fig_1}
\end{figure*}

We start with a global filter \cite{gu2022efficiently} as our baseline operator whose kernel size is equal to the sequence length after transforming latent representation into the frequency domain via Fourier transform, as shown in Figure \ref{fig_1a}. The operator can efficiently modeling global and cyclic information through the simple element-wise multiplication which mathematically equals to circular global convolution that can be served as a global token mixer \cite{Wei2022ActiveTM} in the origin latent space \cite{McGillem1984ContinuousAD} with only $\mathcal{O}(N\log N)$ complexity. Although this operator is able to handle the complex long-range arena (LRA) sequence \cite{Tay2020LongRA}, language \cite{Li2022WhatMC} and image \cite{rao2021global} data, it fall short of the state-of-the-art methods for time series analysis \cite{Zhou2022FiLMFI}. The reason behind can be explained by lacking key properties for time series spectrum feature which should be introduced to the designation of the global kernel.

Consider a time series data $\mathbb{R}^{m\times L}$ with $m$ variates and length $L$. We can project it to a frequency domain feature $\mathbb{C}^{m \times C \times N}$ under the channel independent setting \cite{nie2022time} via Fourier transform, where $C$ is the channel number, $N=[\frac{L}{2}+1]$ is the frequency length. As a result, there are three dimensions to consider for modeling the time series spectrum feature, i.e. the variable dimension, the channel dimension, the frequency dimension. We will detailed the insight of our design follow the three perspectives as shown in Figure \ref{fig_1}.

\begin{figure}[h]
	\begin{center}
		\includegraphics[width=\linewidth]{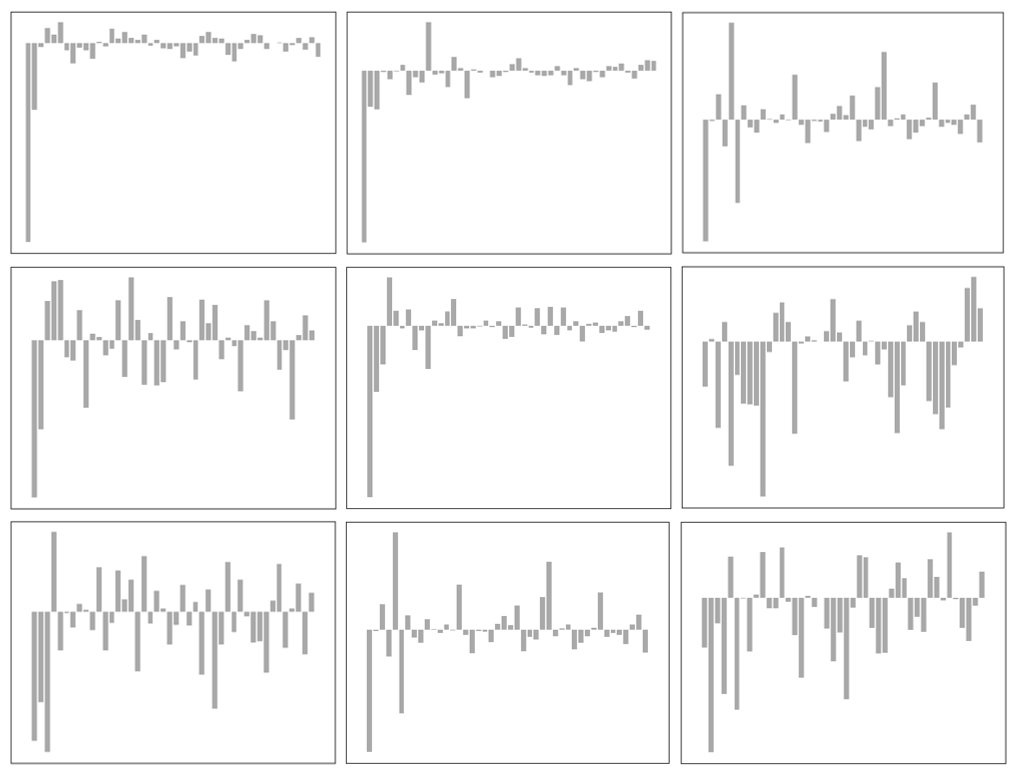}
	\end{center}
	\caption{Visualization of channel redundancy in the frequency domain. Only the real part is visualized. Qualitatively, we can see the high redundancies across different channels.}
	\label{fig_vis_redundant}
\end{figure}

The first one is the channel redundancy property among the channel dimension, i.e. the feature map in the frequency domain shares high similarities among different channels as shown in Figure \ref{fig_vis_redundant}, where we randomly visualize a sequence with input lenght $720$ on the latent frequency space on ETTh1 dataset. Although this redundant information often guarantees a comprehensive understanding of the input data which cannot be directly removed \cite{Han2019GhostNetMF,Nie2021GhostSRLG}, the collapsed feature still affects the expressive power significantly \cite{Wang2023PanGupiEL}. To make full use of the channel redundancy property as well as increasing channel-wise feature diversity, we propose a simple yet effective partial operation which apply the global filter on only a part of the input channels for frequency feature extraction and leaves the remaining channels untouched as illustrated in Figure \ref{fig_1b}. 

Secondly we analyze the frequency dimension. Similiar to images \cite{Ruderman1994TheSO,riad2022diffstride}, sounds \cite{Singh2003ModulationSO}, and surfaces \cite{Kuroki2018HapticTP}, energy of time series signals is highly distributed in the lower frequencies. However, time series energy distributed in the higher parts plays a key role as discussed in \cite{wu2023timesnet}, which reflects important trends \cite{zhou2022fedformer}. Inspired by the sparse kernel \cite{dai2022revisiting} for extracting the most informative elements, we let the kernel operate on the salient frequency parts only as shown in Figure \ref{fig_1c}, while leaving other low signal-to-noise ratio (SNR) parts unchanged which are still informative even less important. In this way, the model can preserve critical historical information while avoiding overfitting to noise. 

Thirdly, the adaptability to semantics also matters under the channel independence setting \cite{nie2022time,Zeng2022AreTE} among the variable dimension. The feature extractor should be variable semantic adaptive not only instance-aware \cite{NIPS2017_3f5ee243,Xiao2022DynamicSN} as shown in Figure \ref{fig_1d}. For example, the Traffic dataset in the forecasting task has $862$ independent variable dimensions. Besides, the spectrum feature is not translation equivariance \cite{Craig1985A} which requires the operator to be spatial-specific and can adaptively allocate the weights over different frequency bands. Thus, instead of using a static filter, we adopt an simple linear layer to learn semantic adaptive filter from the frequency representations. Then we calculate the element-wise multiplication between the learned filter and the spectrum feature.

Empowering by above key insights, our proposed module can dynamically extract the most informative parts among the frequency domain gracefully. Then we transform the filtered latent feature back into the original time domain space via inverse Fourier transform and process it with Feed-forward network (FFN) follow the modern architecture design \cite{Wang2023RepViTRM,Liu2022ACF}. 

With the above module as the primary operator, we build our model named as Omni-Dimensional Frequency Learner (ODFL). Empirically, ODFL improves the baseline global kernel significantly as well as achieves the consistent state-of-the-art on five mainstream time series analysis tasks, which demonstrate our consistent superiority and the excellent task-generalization ability.

\begin{figure*}[t]
	\begin{center}
		\includegraphics[width=\linewidth]{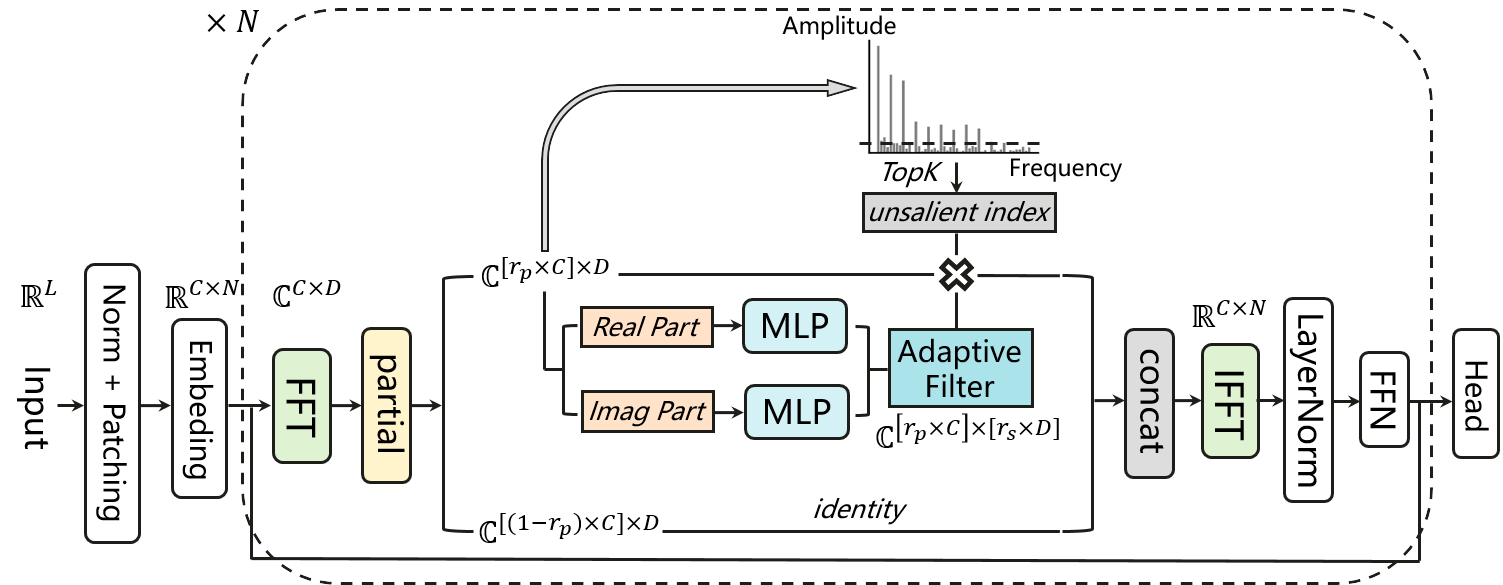}
	\end{center}
	\caption{The architecture of our proposed method. The transformed frequency feature is firstly partial to two parts. Then, our learned adaptive filter is applied on only a few inputs channels and the unsalient frequency bands while leaving the remaining ones untouched. For simplicity, we only take one variable example for visualization.}
	\label{fig:method}
\end{figure*}

\section{Related Work}
Many task-specific methods are proposed recently, including the Transformer- \cite{xu2021anomaly,performer,haoyietal-informer-2021,2019Enhancing,wu2021autoformer,Liu2022NonstationaryTR}, MLP- \cite{2021A,Zhang2022LessIM,oreshkin2019n,challu2022n}, CNN-based \cite{Wang2023MICNML,Liu2021SCINetTS,2017Conditional,Bai2018AnEE,He2019TemporalCN}, and classical \cite{Chen2016XGBoostAS,Taylor2018Forecasting,Berndt1994UsingDT,Rasheed2014AFF,Vlachos2005OnPD} methods. These methods have learned the dependencies in the
time domain. For instance, MICN \cite{Wang2023MICNML} proposes a multi-scale convolution structure to combine local features and global correlations in time series. PatchTST \cite{nie2022time} propose channel-independent learning and patching design for time series forecasting and shows superior performance than linear model \cite{Zeng2022AreTE}. SCINet \cite{Liu2021SCINetTS} introduces a recursive downsample-convolve-interact architecture to model time series with complex temporal dynamics. Besides, DLinear \cite{Zeng2022AreTE} proposes a set of simple one-layer linear model to learn temporal relationships between input and
output sequences. 

Several recent works try to enhance time series feature in the frequency domain for time series forecasting. For example, FEDFormer \cite{zhou2022fedformer} enhances the seasonal-trend decomposition by using self-attention within the frequency domain. S4 \cite{gu2022efficiently,Li2022WhatMC} comes up with recursive memory design for data representation on Long Range Arena. SFM \cite{Zhang2017StockPP} decomposes
the hidden state of LSTM into frequencies. StemGNN \cite{Cao2020SpectralTG} performs graph convolutions based on Graph Fourier Transform (GFT) and computes series correlations based on Discrete Fourier Transform. AutoFormer \cite{wu2021autoformer} replaces self-attention by proposing the auto-correlation mechanism implemented with Fast Fourier Transform. FreTS \cite{Yi2023FrequencydomainMA} propose to directly apply MLPs in the frequency domain to capture global and periodic patterns. FiLM \cite{Zhou2022FiLMFI} applies Legendre Polynomials projections to approximate historical information and uses Fourier projection to remove noise. However, they still fail to serve as a general-purpose model for diverse tasks as well as achieve better performance than traditional time domain methods. We propose a new architecture in the frequency domain that consider the key properties among the variable, channel, frequency dimensions thus can be generalized to various time series task.

To achieve task-general ability, TimesNet \cite{wu2023timesnet} proposes to transform the 1D time-series data into 2D space based on multiple periods to model intraperiod- and interperiod-variations, but it neglects a lot of valuable information in the frequency domain and is heavy-weight. OFA \cite{zhou2023one} explores to use large scale pre-trained frozen language models to serve as an universal compute engine \cite{Lu2022FrozenPT,Giannou2023LoopedTA} to leverage inductive biases learned from language data \cite{Radford2019LanguageMA} instead of introducing inductive biases by architecture designation. Also, Lag-Llama \cite{Rasul2023LagLlamaTF} investigate the use of pre-trained time series models from Monash dataset \cite{Godahewa2021MonashTS} for zero-shot forecasting. Our work takes a step further to enhance the line of deep models and frequency domain feature modeling by introducing general inductive biases to fully leverage the property the time series feature in both time and frequency domain.

\section{Method}
\subsection{Preliminaries: Discrete Fourier Transform}
Given a sequence of time series data $x[n], 0\le n\le N-1$, Discrete Fourier Transform (DFT) converts the sequence into the frequency domain by:

\begin{equation}
	\label{equ:dft}
	X[k]=\sum_{n=0}^{N-1} x[n] e^{-j(2 \pi / N) k n}, 0\le k\le N-1.
\end{equation}

where $j$ is the imaginary unit. $X[k]$ represents the spectrum of the sequence $x[n]$ at the frequency $\omega_k = \frac{2\pi k}{N}$. Conversely, we can recover the original sequence by the inverse DFT (IDFT) given the $X[k]$:

\begin{equation}
	x[n] = \frac{1}{N} \sum_{k=0}^{N - 1}X[k]e^{j(2\pi/N)kn}.
	\label{equ:idft}
\end{equation}

For real input $x[n]$, DFT is conjugate symmetric, i.e., $X[N - k] =X[k]$ (see Appendix \ref{proof:real_signals}). Thus, we can only keep half of the DFT $\{X[k]: 0\le k\le \lceil \frac{N}{2} \rceil\}$ to save memory and computation costs. To compute the DFT, the Fast Fourier Transform (FFT) is widely used which reduces the complexity of DFT from $\mathcal{O}(N^2)$ to $\mathcal{O}(N\log N)$. Inverse fast Fourier transform (IFFT) can also be used for computing IDFT, which is similar to FFT.

\subsection{Frequency Learner Architecture}
\label{sec:method}

\paragraph{Overall Structure.} The overall architecture we employ is depicted in Figure \ref{fig:method}. Given the length L input time series data $\bm{x}\in \mathbb{R}^{m\times L}$ with $m$ variates normalized by ReVIN instance normalization \cite{Kim2022ReversibleIN}, we follow \cite{nie2022time} to split $\bm{x}$ to $m$ univariate series $\bm{x}^{(i)}\in \mathbb{R}^{1\times L}, i=1,...,m$, where each of them is fed independently into the ODFL model. Each input univariate series $\bm{x}^{(i)}$ is divided into patches with patch length $P$ and stride $S$ and generate a sequence $\bm{x}^{(i)}_{p}\in \mathbb{R}^{P\times N}$ where $N=\left\lfloor\frac{(L-P)}{S}\right\rfloor+2$ is the patches number. Then the sequence is project to a sequence feature $\bm{x}^{(i)}\in \mathbb{R}^{C\times N}$ by a linear layer. Finally, the feature is fed into our ODFL blocks and corresponding head to obtain the final output.

\paragraph{Baseline.} We start with a simple global filter to process the feature $\bm{x}\in \mathbb{R}^{C\times N}$ as our baseline, index $i$ is omit for simplicity. We firstly transform $\bm{x}$ into the frequency domain via FFT. Since $\bm{x}$ is a real tensor, the frequency domain feature is conjugate symmetric. Thus, we can only take half of the values but preserve the complete information (Appendix \ref{proof:real_signals}) and get the complex frequency domain feature $\bm{X}\in \mathbb{C}^{C\times D}$, where $D=[\frac{N}{2}+1]$ is the frequency length. Then we extract feature by element-wise multiplying a learnable filter:

\begin{equation}
	\label{equ:baseline}
	\bm{\widetilde{X}} = \bm{X} \odot \bm{K}.
\end{equation}

where $\bm{K} \in\mathbb{C}^{C \times D}$. The equation is equivalent to the global circular convolution with global kernel in the time domain which can serve as a global token mixer (Appendix \ref{proof:circular_conv}).

\paragraph{Extract Feature on Partial Channels.} While the redundant features \cite{Han2019GhostNetMF} in $\bm{X}\in \mathbb{C}^{C\times D}$ provide a wealth of information, they decrease the diversity of the filter as well as increasing computation costs. 

To takes advantage of the redundancy, we apply the filter on only a part of the spectrum feature channels while leaving the remaining ones untouched: 

\begin{equation}
	\label{equ:partial}
	\bm{\widetilde{X}}_{1: [r_{p}\times D]} = \bm{X}_{1: [r_{p}\times D]} \odot \bm{K}.
\end{equation}

where $r_{p}$ is the partial ratio (PR), $\bm{K} \in \mathbb{C}^{[r_{p}\times C] \times D}$. 

We calculate the effective dimension ratio $r_{d(\epsilon)}$ follow the analysis method from \cite{Cai2021IsotropyIT}. The effective dimension $d(\epsilon)$ is defined as the minimum principal component numbers that occupy the explained variance ratio of $\epsilon$ in a principal component analysis (PCA), and the effective dimension ratio $r_{d(\epsilon)}$ is the proportion to dimension number. Thus, a more powerful representation of a feature among the channel dimension would result in a larger effective dimension ratio and vice versa. 

\begin{figure}[h]
	\begin{center}
		\includegraphics[width=\linewidth]{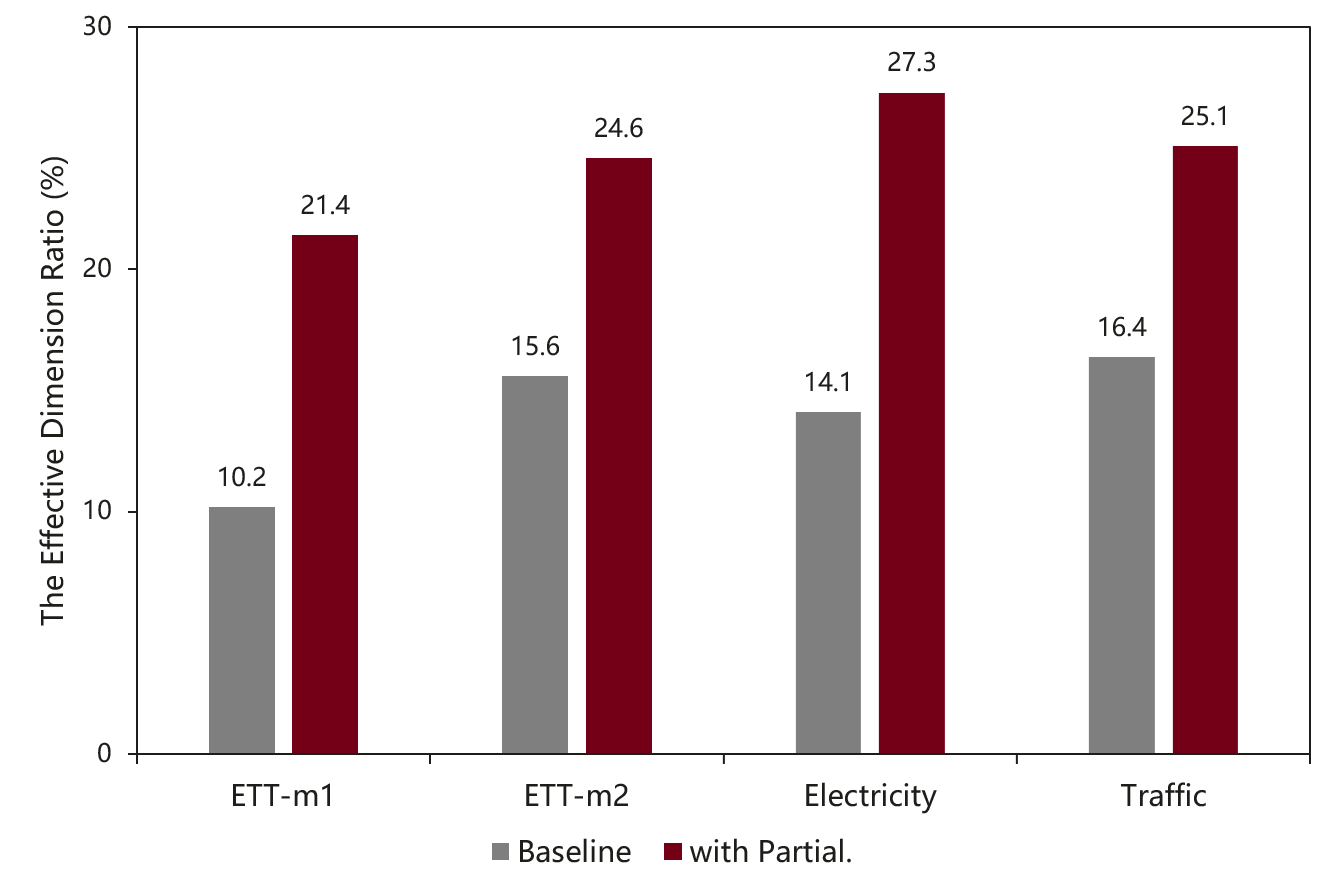}
	\end{center}
	\vspace{-8pt}
	\caption{The effective dimension ratio $r_{d(0.8)}$ of our model which reflects the feature redundant and diversity. Higher effective dimension ratio indicates more diverse feature among channel dimension.}
	\label{fig_redundant}
\end{figure}

The smaller ratio of the baseline model in Figure \ref{fig_redundant} shows the feature collapse problem. 
It can be observed that adding the partial operation increase the effective dimension to the great extent, indicating the significant role of our proposed components in channel-wise feature diversity, by separating the latent feature into two parts and transforming the processed one into another feature space.

\paragraph{Extract Feature in Un-Salient Parts.} Consider the sparsity \cite{wu2023timesnet} and noisy \cite{1980The} in the frequency domain, we take a step further to consider the salient spectrum parts only which can serve as the low-rank approximation for fourier basis \cite{zhou2022fedformer}. We first get the amplitude of each frequency which represents its importance as shown in the upper part in Figure \ref{fig:method}:

\begin{equation}
	\mathbf{A}=\operatorname{Avg}\left(\operatorname{Amp}\left(\bm{X}\right)\right).
\end{equation}

where $\operatorname{Amp}(\cdot)$ denote the calculation of amplitude values. $\mathbf{A}\in\mathbb{R}^{D}$ represents the calculated amplitude of each frequency, which is averaged from $C$ dimensions by $\operatorname{Avg}(\cdot)$. Then we select the tok-k amplitude frequencies as the salient frequencies:

\begin{equation}
	\label{equ:salient_frequency}
	\left\{f_1, \cdots, f_k\right\}=
	\underset{f_* \in\left\{1, \cdots, D\right\}}{\arg \operatorname{Topk}}(\mathbf{A}).
\end{equation}

where $k$ is a hyper-parameter, $k \leq D$, and $f_1 < \cdots < f_k$. We denote $r_{s} = \frac{k}{D}$ as the sparsity ratio (SR)\label{equ: sr}. Afterward, we can extract features from the salient bands by multiplying our learnable sparse kernel: 

\begin{equation}
	\label{equ:salient}
	\bm{\widetilde{X}}_{1: [r_{p}\times D]}\left[ f_1, \cdots, f_k \right] = \bm{X}_{1: [r_{p}\times D]}\left[ f_1, \cdots, f_k \right] \odot \bm{K}.
\end{equation}

where $\bm{K} \in\mathbb{C}^{[r_{p} \times C] \times [r_{s} \times D]}$. Here for other un-salient frequencies $f \notin \left\{f_1, \cdots, f_k\right\}$, we simply keep them unchanged in $\bm{\widetilde{X}}$ since these low SNR parts still have valuable information when transforming back into the time domain, and dropping them (i.e. setting the values to zero) will cause the Picket Fence Effect \cite{2008Eliminating}. 

\paragraph{Learn Semantic-Adaptive Filter Among the Variable Dimension.} The adaptability to semantics has played the key role in feature modeling \cite{Wei2022ActiveTM,Jie2019Squeeze} instead of passively determining by manual designed rules \cite{Tolstikhin2021MLPMixerAA} especially under the channel independence setting. 
With the expectation for our operator as semantic-adaptive, the weights $\bm{K}$ should be adaptive to $\bm{X}$. 
We utilize simply linear layer by separately considering the real and imaginary mappings to learn the adaptive filter motivated by dynamic convolution \cite{Chen2019DynamicCA}. The learned respective real/imaginary parts are then stacked to recover the frequency components in order to obtain the adaptive filter $\bm{K} \in\mathbb{C}^{[r_{p} \times C] \times [r_{s} \times D]}$ in Equation (\ref{equ:salient}).

\paragraph{Output.} The feature is transform back into the original time domain by using IFFT algorithm. Then, LayerNorm \cite{2016Layer} and inverted FFN layer \cite{Sandler2018MobileNetV2IR} is applied to fully leverage the information from all channels. We stack multiple proposed blocks for constructing our Frequency Learner model. Finally, the output time domain feature from the last module is passed to corresponding head according to different tasks.

\begin{table*}[htbp]
	\caption{Summary of experiment benchmarks. See detailed benchmark description in Appendix \ref{appen:datasets}.}\label{tab:benchmarks}
	\centering
	\begin{threeparttable}
		\renewcommand{\multirowsetup}{\centering}
		\setlength{\tabcolsep}{2pt}
		\begin{tabular}{c|l|c|c}
			\toprule
			{Tasks} & {Benchmarks} & {Metrics} & {Series Length} \\
			\toprule
			{\multirow{3}{*}{Forecasting}}  & {Long-term: ETT (4 subsets), Electricity,} & {\multirow{2}{*}{MSE, MAE}} & {96$\sim$720} \\
			& {Traffic, Weather, ILI} & & {(ILI: 24$\sim$60)}\\
			\cmidrule{2-4}
			& {Short-term: M4 (6 subsets)}  & {SMAPE, MASE, OWA} & {6$\sim$48} \\
			\midrule
			{Imputation} & {ETT (4 subsets), Electricity, Weather} & {MSE, MAE} & 96 \\
			\midrule
			{Classification} & {UEA (10 subsets)} & {Accuracy} & {29$\sim$1751} \\
			\midrule
			{Anomaly Detection} & {SMD, MSL, SMAP, SWaT, PSM} & {Precision, Recall, F1-Socre} & 100 \\
			\bottomrule
		\end{tabular}
	\end{threeparttable}
\end{table*}

\begin{table*}[ht]
	\caption{Long-term forecasting task. All the results are averaged from 4 different prediction lengths, that is $\{24,36,48,60\}$ for ILI and $\{96,192,336,720\}$ for the others. See Table \ref{tab:full_forecasting_results} in Appendix \ref{appendix:full} for the full results. The best and second best results are in \textbf{Bold} and \underline{underline}.}
	\label{tab:long_term_forecasting_results}
	\centering
	\begin{small}
	\begin{threeparttable}
			\renewcommand{\multirowsetup}{\centering}
			\setlength{\tabcolsep}{0.5pt}
			\begin{tabular}{c|cc|cc|cc|cc|cc|cc|cc|cc|cc|cc|cc|cc}
				\toprule
				\multicolumn{1}{c}{\multirow{2}{*}{Models}} & 
				\multicolumn{2}{c}{\rotatebox{0}{\scalebox{1}{ODFL}}} &
				\multicolumn{2}{c}{\rotatebox{0}{\scalebox{1}{TimesNet}}} &
				\multicolumn{2}{c}{\rotatebox{0}{\scalebox{1}{PatchTST}}} &
				\multicolumn{2}{c}{\rotatebox{0}{\scalebox{1}{Cross.}}} &
				\multicolumn{2}{c}{\rotatebox{0}{\scalebox{1}{DLinear}}} &
				\multicolumn{2}{c}{\rotatebox{0}{\scalebox{1}{FED.}}} & \multicolumn{2}{c}{\rotatebox{0}{\scalebox{1}{MICN}}} & \multicolumn{2}{c}{\rotatebox{0}{\scalebox{1}{SCINet}}} & \multicolumn{2}{c}{\rotatebox{0}{\scalebox{1}{ETS.}}} &  \multicolumn{2}{c}{\rotatebox{0}{\scalebox{1}{OFA}}} & \multicolumn{2}{c}{\rotatebox{0}{\scalebox{1}{LightTS}}} & \multicolumn{2}{c}{\rotatebox{0}{\scalebox{1}{FreTS}}}
				\\
				\multicolumn{1}{c}{}&
				\multicolumn{2}{c}{\scalebox{1}{(Ours)}} & \multicolumn{2}{c}{\scalebox{1}{(2023)}} & 
				\multicolumn{2}{c}{\scalebox{1}{(2023)}} &
				\multicolumn{2}{c}{\scalebox{1}{(2023)}} &
				\multicolumn{2}{c}{\scalebox{1}{(2023)}} & \multicolumn{2}{c}{\scalebox{1}{(2022)}} & \multicolumn{2}{c}{\scalebox{1}{(2023)}} & \multicolumn{2}{c}{\scalebox{1}{(2022)}} & \multicolumn{2}{c}{\scalebox{1}{(2022)}} &  \multicolumn{2}{c}{\scalebox{1}{(2023)}} &  \multicolumn{2}{c}{\scalebox{1}{(2022)}} &  \multicolumn{2}{c}{\scalebox{1}{(2023)}} 
				\\
				\cmidrule(lr){0-0} \cmidrule(lr){2-3} \cmidrule(lr){4-5}\cmidrule(lr){6-7} \cmidrule(lr){8-9}\cmidrule(lr){10-11}\cmidrule(lr){12-13}\cmidrule(lr){14-15}\cmidrule(lr){16-17}\cmidrule(lr){18-19}\cmidrule(lr){20-21}\cmidrule(lr){22-23}\cmidrule(lr){24-25}
				\multicolumn{1}{c}{\scalebox{1}{Metric}} & \scalebox{1}{MSE} & \scalebox{1}{MAE} & \scalebox{1}{MSE} & \scalebox{1}{MAE} & \scalebox{1}{MSE} & \scalebox{1}{MAE} & \scalebox{1}{MSE} & \scalebox{1}{MAE} & \scalebox{1}{MSE} & \scalebox{1}{MAE} & \scalebox{1}{MSE} & \scalebox{1}{MAE} & \scalebox{1}{MSE} & \scalebox{1}{MAE} & \scalebox{1}{MSE} & \scalebox{1}{MAE} & \scalebox{1}{MSE} & \scalebox{1}{MAE} & \scalebox{1}{MSE} & \scalebox{1}{MAE} & \scalebox{1}{MSE} & \scalebox{1}{MAE} & \scalebox{1}{MSE} & \scalebox{1}{MAE} \\
				\toprule
				\scalebox{1}{ETTm1} &\textbf{\scalebox{1}{0.343}} &\textbf{\scalebox{1}{0.378}}  &\scalebox{1}{0.400} &\scalebox{1}{0.406} & \underline{\scalebox{1}{0.351}} & \scalebox{1}{0.381} & \scalebox{1}{0.431} &\scalebox{1}{0.443} &\scalebox{1}{0.357} &\underline{\scalebox{1}{0.379}} &\scalebox{1}{0.382} &\scalebox{1}{0.422} &\scalebox{1}{0.383} &\scalebox{1}{0.406} &\scalebox{1}{0.387} &\scalebox{1}{0.411} &\scalebox{1}{0.429} &\scalebox{1}{0.425} &\scalebox{1}{0.352} &\scalebox{1}{0.383} &\scalebox{1}{0.435} &\scalebox{1}{0.437}
				&\scalebox{1}{0.362} 
				&\scalebox{1}{0.386}\\
				\midrule
				\scalebox{1}{ETTm2} &\textbf{\scalebox{1}{0.246}} &\textbf{\scalebox{1}{0.313}}  &{\scalebox{1}{0.291}} &{\scalebox{1}{0.333}} &\underline{\scalebox{1}{0.255}} & \underline{\scalebox{1}{0.315}} & \scalebox{1}{0.632} &\scalebox{1}{0.578} &\scalebox{1}{0.267} &\scalebox{1}{0.332} &\scalebox{1}{0.292} &\scalebox{1}{0.343} &\scalebox{1}{0.277} &\scalebox{1}{0.336} &\scalebox{1}{0.294} &\scalebox{1}{0.355} &\scalebox{1}{0.293} &\scalebox{1}{0.342} &\scalebox{1}{0.266} &\scalebox{1}{0.326} &\scalebox{1}{0.409} &\scalebox{1}{0.436}
				&\scalebox{1}{0.274} 
				&\scalebox{1}{0.335}\\
				\midrule
				\scalebox{1}{ETTh1} &\textbf{\scalebox{1}{0.413}} &\underline{\scalebox{1}{0.432}}  &\scalebox{1}{0.458} & {\scalebox{1}{0.450}} &\textbf{\scalebox{1}{0.413}} & \underline{\scalebox{1}{0.432}} & \scalebox{1}{0.441} &\scalebox{1}{0.465} &\scalebox{1}{0.423} &\scalebox{1}{0.437} &{\scalebox{1}{0.428}} &\scalebox{1}{0.454} &\scalebox{1}{0.433} &\scalebox{1}{0.462} &\scalebox{1}{0.460} &\scalebox{1}{0.462} &\scalebox{1}{0.542} &\scalebox{1}{0.510} &\scalebox{1}{0.427} &\textbf{\scalebox{1}{0.426}} &\scalebox{1}{0.491} &\scalebox{1}{0.479}
				&\scalebox{1}{0.426} 
				&\scalebox{1}{0.448}\\
				\midrule
				\scalebox{1}{ETTh2} &\underline{\scalebox{1}{0.331}} &\underline{\scalebox{1}{0.384}} &\scalebox{1}{0.414} &{\scalebox{1}{0.427}} &\textbf{\scalebox{1}{0.330}} & \textbf{\scalebox{1}{0.380}} &\scalebox{1}{0.835} &\scalebox{1}{0.676} &\scalebox{1}{0.431} &\scalebox{1}{0.447} &\scalebox{1}{0.388} &\scalebox{1}{0.434} &\scalebox{1}{0.385} &\scalebox{1}{0.430} &\scalebox{1}{0.371} &\scalebox{1}{0.410} &\scalebox{1}{0.439} &\scalebox{1}{0.452} &\scalebox{1}{0.346} &\scalebox{1}{0.394} &\scalebox{1}{0.602} &\scalebox{1}{0.543}
				&\scalebox{1}{0.413} 
				&\scalebox{1}{0.435}\\
				\midrule
				\scalebox{1}{Electricity} &\textbf{\scalebox{1}{0.158}} &\underline{\scalebox{1}{0.253}}  &\scalebox{1}{0.192} &\scalebox{1}{0.295} &\underline{\scalebox{1}{0.161}} &\textbf{\scalebox{1}{0.252}} & \scalebox{1}{0.278} &\scalebox{1}{0.339} &\scalebox{1}{0.177} &\scalebox{1}{0.274} &\scalebox{1}{0.207} &\scalebox{1}{0.321} &\scalebox{1}{0.182} &\scalebox{1}{0.292} &\scalebox{1}{0.191} &\scalebox{1}{0.277} &\scalebox{1}{0.208} &\scalebox{1}{0.323} &\scalebox{1}{0.167} &\scalebox{1}{0.263} &\scalebox{1}{0.229} &\scalebox{1}{0.329}
				&\scalebox{1}{0.164} 
				&\scalebox{1}{0.264}\\
				\midrule
				\scalebox{1}{Traffic} &\textbf{\scalebox{1}{0.388}} &\underline{\scalebox{1}{0.265}}  &\scalebox{1}{0.620} &\scalebox{1}{0.336} &\underline{\scalebox{1}{0.390}} & \textbf{\scalebox{1}{0.263}} &\scalebox{1}{0.513} &\scalebox{1}{0.290} &\scalebox{1}{0.434} &\scalebox{1}{0.295} &\scalebox{1}{0.604} &\scalebox{1}{0.372} &\scalebox{1}{0.525} &\scalebox{1}{0.312} &\scalebox{1}{0.577} &\scalebox{1}{0.376} &\scalebox{1}{0.621} &\scalebox{1}{0.396} &\scalebox{1}{0.414} &\scalebox{1}{0.294} &\scalebox{1}{0.622} &\scalebox{1}{0.392}
				&\scalebox{1}{0.423} 
				&\scalebox{1}{0.292}\\
				\midrule
				\scalebox{1}{Weather} &\textbf{\scalebox{1}{0.221}} &\textbf{\scalebox{1}{0.260}} &\scalebox{1}{0.259} &\scalebox{1}{0.287} &\underline{\scalebox{1}{0.225}} &\underline{\scalebox{1}{0.264}} &\scalebox{1}{0.230} &\scalebox{1}{0.290} &\scalebox{1}{0.240} &\scalebox{1}{0.300} &\scalebox{1}{0.310} &\scalebox{1}{0.357} &\scalebox{1}{0.242} &\scalebox{1}{0.298} &\scalebox{1}{0.287} &\scalebox{1}{0.317} &\scalebox{1}{0.271} &\scalebox{1}{0.334} &\scalebox{1}{0.237} &\scalebox{1}{0.270} &\scalebox{1}{0.261} &\scalebox{1}{0.312}
				&\scalebox{1}{0.232} 
				&\scalebox{1}{0.276}\\
				\midrule
				\scalebox{1}{ILI} & \textbf{\scalebox{1}{1.431}} &\textbf{\scalebox{1}{0.792}} &\scalebox{1}{2.139} &\scalebox{1}{0.931} &\underline{\scalebox{1}{1.443}} &\underline{\scalebox{1}{0.797}} & \scalebox{1}{3.361} &\scalebox{1}{1.235} &\scalebox{1}{2.160} &\scalebox{1}{1.041} &\scalebox{1}{2.597} &\scalebox{1}{1.070} &\scalebox{1}{2.567} &\scalebox{1}{1.055} &\scalebox{1}{2.253} &\scalebox{1}{1.021} &\scalebox{1}{2.497} &\scalebox{1}{1.004} &\scalebox{1}{1.925} &\scalebox{1}{0.903} &\scalebox{1}{7.382} &\scalebox{1}{2.003}
				&\scalebox{1}{1.976} 
				&\scalebox{1}{0.972}\\
				\bottomrule
			\end{tabular}
	\end{threeparttable}
\end{small}
\end{table*}

\begin{table*}[ht]
	\caption{Short-term forecasting task on M4. The prediction lengths are in $[6,48]$, and results are weighted averaged from several datasets under different sample intervals. See Table \ref{tab:full_forecasting_results_m4} in Appendix \ref{appendix:full} for full results.}
	\label{tab:short_term_forecasting_results}
	\centering
	\begin{threeparttable}
			\renewcommand{\multirowsetup}{\centering}
			\setlength{\tabcolsep}{0.2pt}
			\begin{tabular}{c|ccccccccccccccccccccc}
				\toprule
				\multicolumn{1}{c}{\multirow{2}{*}{Models}} & 
				\multicolumn{1}{c}{\rotatebox{0}{\scalebox{1}{ODFL}}} &
				\multicolumn{1}{c}{\rotatebox{0}{\scalebox{1}{TimesNet}}} &
				\multicolumn{1}{c}{\rotatebox{0}{\scalebox{1}{PatchTST}}} &
				\multicolumn{1}{c}{\rotatebox{0}{\scalebox{1}{Cross.}}} &
				\multicolumn{1}{c}{\rotatebox{0}{\scalebox{1}{DLinear}}} &
				\multicolumn{1}{c}{\rotatebox{0}{\scalebox{1}{FED.}}} &
				\multicolumn{1}{c}{\rotatebox{0}{\scalebox{1}{MICN}}} &
				\multicolumn{1}{c}{\rotatebox{0}{\scalebox{1}{SCINet}}} & \multicolumn{1}{c}{\rotatebox{0}{\scalebox{1}{ETS.}}} & \multicolumn{1}{c}{\rotatebox{0}{\scalebox{1}{OFA}}} & \multicolumn{1}{c}{\rotatebox{0}{\scalebox{1}{LightTS}}} & \multicolumn{1}{c}{\rotatebox{0}{\scalebox{1}{Station.}}} &  \multicolumn{1}{c}{\rotatebox{0}{\scalebox{1}{N-HiTS}}} & \multicolumn{1}{c}{\rotatebox{0}{\scalebox{1}{N-BEATS}}} & \multicolumn{1}{c}{\rotatebox{0}{\scalebox{1}{FreTS}}}\\
				\multicolumn{1}{c}{ } & 
				\multicolumn{1}{c}{\scalebox{1}{(Ours)}} & 
				\multicolumn{1}{c}{\scalebox{1}{(2023)}} &
				\multicolumn{1}{c}{\scalebox{1}{(2023)}} &
				\multicolumn{1}{c}{\scalebox{1}{(2023)}} &
				\multicolumn{1}{c}{\scalebox{1}{(2023)}} &
				\multicolumn{1}{c}{\scalebox{1}{(2022)}} &
				\multicolumn{1}{c}{\scalebox{1}{(2023)}} & \multicolumn{1}{c}{\scalebox{1}{(2022)}} & \multicolumn{1}{c}{\scalebox{1}{(2022)}} & \multicolumn{1}{c}{\scalebox{1}{(2023)}} & \multicolumn{1}{c}{\scalebox{1}{(2022)}} & \multicolumn{1}{c}{\scalebox{1}{(2022)}} & \multicolumn{1}{c}{\scalebox{1}{(2022)}} & \multicolumn{1}{c}{\scalebox{1}{(2019)}} & \multicolumn{1}{c}{\scalebox{1}{(2023)}}\\
				\toprule
				\scalebox{1}{SMAPE} &\textbf{\scalebox{1}{11.734}} &\underline{\scalebox{1}{11.829}} &\scalebox{1}{12.059} &\scalebox{1}{13.474} & \scalebox{1}{13.639} &\scalebox{1}{12.840} &\scalebox{1}{13.130} &\scalebox{1}{12.396} &\scalebox{1}{14.718} &\scalebox{1}{11.991} &\scalebox{1}{13.525} &\scalebox{1}{12.780}  &\scalebox{1}{11.927}  &\scalebox{1}{11.851} &\scalebox{1}{11.905} 
				\\
				\scalebox{1}{MASE} &\textbf{\scalebox{1}{1.576}} &\underline{\scalebox{1}{1.585}} &\scalebox{1}{1.623} &\scalebox{1}{1.866} & \scalebox{1}{2.095} &\scalebox{1}{1.701} &\scalebox{1}{1.896} &\scalebox{1}{1.677} &\scalebox{1}{2.408} &\scalebox{1}{1.600} &\scalebox{1}{2.111}  &\scalebox{1}{1.756}  &\scalebox{1}{1.613}  &\scalebox{1}{1.599} &\scalebox{1}{1.605} 
				\\
				\scalebox{1}{OWA} &\textbf{\scalebox{1}{0.845}}  &\underline{\scalebox{1}{0.851}} &\scalebox{1}{0.869} &\scalebox{1}{0.985} &\scalebox{1}{1.051} &\scalebox{1}{0.918} &\scalebox{1}{0.980} &\scalebox{1}{0.894} &\scalebox{1}{1.172} &\scalebox{1}{0.861} &\scalebox{1}{1.051} &\scalebox{1}{0.930}  &\scalebox{1}{0.861}  &\scalebox{1}{0.855}  &\scalebox{1}{0.858} 
				\\
				\bottomrule
			\end{tabular}
	\end{threeparttable}
\end{table*}

\section{Experiments}
\label{sec:exp}
In this section, we evaluate the performance of our proposed method on five mainstream tasks, including short- and long-term forecasting, imputation, classification and anomaly detection to verify its generality. Then, we provide an ablation study for the impact of each choice and analyze the model properties.

\subsection{Benchmarks.} 
The summary of benchmarks adhered to the experimental settings of TimesNet \cite{wu2023timesnet} are list in Table \ref{tab:benchmarks}. Follow OFA \cite{zhou2023one} and PatchTST \cite{nie2022time}, Exchange dataset is not contained for the long-term forecasting task, since \cite{Zeng2022AreTE} shows that simply repeating the last value in the look-back window can outperform or be comparable to the best results. 

\begin{table*}[htp]
	\caption{Imputation task. We randomly mask $\{12.5\%, 25\%, 37.5\%, 50\%\}$ time points in length-96 time series. The results are averaged from 4 different mask ratios. See Table \ref{tab:full_imputation_results} in Appendix \ref{appendix:full} for full results.}
	\label{tab:imputation_results}
	\centering
	\begin{threeparttable}
		\begin{small}
			\renewcommand{\multirowsetup}{\centering}
			\setlength{\tabcolsep}{0.5pt}
			\begin{tabular}{c|cc|cc|cc|cc|cc|cc|cc|cc|cc|cc|cc|cc}
				\toprule
				\multicolumn{1}{c}{\multirow{2}{*}{Models}} & 
				\multicolumn{2}{c}{\rotatebox{0}{\scalebox{1}{ODFL}}} &
				\multicolumn{2}{c}{\rotatebox{0}{\scalebox{1}{TimesNet}}} &
				\multicolumn{2}{c}{\rotatebox{0}{\scalebox{1}{PatchTST}}} &
				\multicolumn{2}{c}{\rotatebox{0}{\scalebox{1}{Cross.}}} &
				\multicolumn{2}{c}{\rotatebox{0}{\scalebox{1}{DLinear}}} &
				\multicolumn{2}{c}{\rotatebox{0}{\scalebox{1}{FED.}}} & \multicolumn{2}{c}{\rotatebox{0}{\scalebox{1}{MICN}}} & \multicolumn{2}{c}{\rotatebox{0}{\scalebox{1}{SCINet}}} & \multicolumn{2}{c}{\rotatebox{0}{\scalebox{1}{ETS.}}} &  \multicolumn{2}{c}{\rotatebox{0}{\scalebox{1}{OFA}}} & \multicolumn{2}{c}{\rotatebox{0}{\scalebox{1}{LightTS}}} & \multicolumn{2}{c}{\rotatebox{0}{\scalebox{1}{FreTS}}} \\
				\multicolumn{1}{c}{} & 
				\multicolumn{2}{c}{\scalebox{1}{(Ours)}} & \multicolumn{2}{c}{\scalebox{1}{(2023)}} & 
				\multicolumn{2}{c}{\scalebox{1}{(2023)}} &
				\multicolumn{2}{c}{\scalebox{1}{(2023)}} &
				\multicolumn{2}{c}{\scalebox{1}{(2023)}} & \multicolumn{2}{c}{\scalebox{1}{(2022)}} & \multicolumn{2}{c}{\scalebox{1}{(2023)}} & \multicolumn{2}{c}{\scalebox{1}{(2022)}} & \multicolumn{2}{c}{\scalebox{1}{(2022)}} &  \multicolumn{2}{c}{\scalebox{1}{(2023)}} & \multicolumn{2}{c}{\scalebox{1}{(2022)}} & \multicolumn{2}{c}{\scalebox{1}{(2023)}} \\
				\cmidrule(lr){0-0} \cmidrule(lr){2-3} \cmidrule(lr){4-5}\cmidrule(lr){6-7} \cmidrule(lr){8-9}\cmidrule(lr){10-11}\cmidrule(lr){12-13}\cmidrule(lr){14-15}\cmidrule(lr){16-17}\cmidrule(lr){18-19}\cmidrule(lr){20-21}\cmidrule(lr){22-23}\cmidrule(lr){24-25}
				\multicolumn{1}{c}{\scalebox{1}{Metric}} & \scalebox{1}{MSE} & \scalebox{1}{MAE} & \scalebox{1}{MSE} & \scalebox{1}{MAE} & \scalebox{1}{MSE} & \scalebox{1}{MAE} & \scalebox{1}{MSE} & \scalebox{1}{MAE} & \scalebox{1}{MSE} & \scalebox{1}{MAE} & \scalebox{1}{MSE} & \scalebox{1}{MAE} & \scalebox{1}{MSE} & \scalebox{1}{MAE} & \scalebox{1}{MSE} & \scalebox{1}{MAE} & \scalebox{1}{MSE} & \scalebox{1}{MAE} & \scalebox{1}{MSE} & \scalebox{1}{MAE} & \scalebox{1}{MSE} & \scalebox{1}{MAE} & \scalebox{1}{MSE} & \scalebox{1}{MAE} \\
				\toprule
				\scalebox{1}{ETTm1} &\textbf{\scalebox{1}{0.016}} &\textbf{\scalebox{1}{0.087}}  &\underline{\scalebox{1}{0.027}} &{\scalebox{1}{0.107}} & \scalebox{1}{0.045} & \scalebox{1}{0.133} &\scalebox{1}{0.041} &\scalebox{1}{0.143} &\scalebox{1}{0.093} &\scalebox{1}{0.206} &\scalebox{1}{0.062} &\scalebox{1}{0.177} &\scalebox{1}{0.070} &\scalebox{1}{0.182} &\scalebox{1}{0.039} &\scalebox{1}{0.129} &\scalebox{1}{0.120} &\scalebox{1}{0.253} &\scalebox{1}{0.028} &\underline{\scalebox{1}{0.105}} &\scalebox{1}{0.104} &\scalebox{1}{0.218} &\scalebox{1}{0.036} &\scalebox{1}{0.105} \\
				\midrule
				\scalebox{1}{ETTm2} &\textbf{\scalebox{1}{0.018}} &\textbf{\scalebox{1}{0.083}} &{\scalebox{1}{0.022}} &{\scalebox{1}{0.088}} &\scalebox{1}{0.028} & \scalebox{1}{0.098} &\scalebox{1}{0.046} &\scalebox{1}{0.149} &\scalebox{1}{0.096} &\scalebox{1}{0.208} &\scalebox{1}{0.101} &\scalebox{1}{0.215} &\scalebox{1}{0.144} &\scalebox{1}{0.249} &\scalebox{1}{0.027} &\scalebox{1}{0.102} &\scalebox{1}{0.208} &\scalebox{1}{0.327} &\underline{\scalebox{1}{0.021}} &\underline{\scalebox{1}{0.084}} &\scalebox{1}{0.046} &\scalebox{1}{0.151} &\scalebox{1}{0.030} &\scalebox{1}{0.104} \\
				\midrule
				\scalebox{1}{ETTh1} &\textbf{\scalebox{1}{0.048}} &\textbf{\scalebox{1}{0.148}} &{\scalebox{1}{0.078}} &{\scalebox{1}{0.187}} &\scalebox{1}{0.133} & \scalebox{1}{0.236} &\scalebox{1}{0.132} &\scalebox{1}{0.251} &\scalebox{1}{0.201} &\scalebox{1}{0.306} &\scalebox{1}{0.117} &\scalebox{1}{0.246} &\scalebox{1}{0.125} &\scalebox{1}{0.250} &\scalebox{1}{0.104} &\scalebox{1}{0.216} &\scalebox{1}{0.202} &\scalebox{1}{0.329} &\underline{\scalebox{1}{0.069}} &\underline{\scalebox{1}{0.173}} &\scalebox{1}{0.284} &\scalebox{1}{0.373} &\scalebox{1}{0.097} &\scalebox{1}{0.210} \\
				\midrule
				\scalebox{1}{ETTh2} &\textbf{\scalebox{1}{0.038}} &\textbf{\scalebox{1}{0.125}} &{\scalebox{1}{0.049}} &{\scalebox{1}{0.146}} &\scalebox{1}{0.066} & \scalebox{1}{0.164} &\scalebox{1}{0.122} &\scalebox{1}{0.240} &\scalebox{1}{0.142} &\scalebox{1}{0.259} &\scalebox{1}{0.163} &\scalebox{1}{0.279} &\scalebox{1}{0.205} &\scalebox{1}{0.307} &\scalebox{1}{0.064} &\scalebox{1}{0.165} &\scalebox{1}{0.367} &\scalebox{1}{0.436} &\underline{\scalebox{1}{0.048}} &\underline{\scalebox{1}{0.141}} &\scalebox{1}{0.119} &\scalebox{1}{0.250} &\scalebox{1}{0.067} &\scalebox{1}{0.166} \\
				\midrule
				\scalebox{1}{Electricity} &\textbf{\scalebox{1}{0.079}} &\textbf{\scalebox{1}{0.195}}  &{\scalebox{1}{0.092}} &{\scalebox{1}{0.210}} &\scalebox{1}{0.091} & \scalebox{1}{0.209} &\scalebox{1}{0.083} &\scalebox{1}{0.199} &\scalebox{1}{0.132} &\scalebox{1}{0.260} &\scalebox{1}{0.130} &\scalebox{1}{0.259} &\scalebox{1}{0.119} &\scalebox{1}{0.247} &\underline{\scalebox{1}{0.086}} &\underline{\scalebox{1}{0.201}} &\scalebox{1}{0.214} &\scalebox{1}{0.339} &\scalebox{1}{0.090} &\scalebox{1}{0.207} &\scalebox{1}{0.131} &\scalebox{1}{0.262} &\scalebox{1}{0.099} &\scalebox{1}{0.218}  \\
				\midrule
				\scalebox{1}{Weather} &\textbf{\scalebox{1}{0.025}} &\textbf{\scalebox{1}{0.047}} &\underline{\scalebox{1}{0.030}} &{\scalebox{1}{0.054}} & \scalebox{1}{0.033} & \scalebox{1}{0.057} &\scalebox{1}{0.036} &\underline{\scalebox{1}{0.050}} &\scalebox{1}{0.052} &\scalebox{1}{0.110} &\scalebox{1}{0.099} &\scalebox{1}{0.203} &\scalebox{1}{0.056} &\scalebox{1}{0.128} &\scalebox{1}{0.031} &\scalebox{1}{0.053} &\scalebox{1}{0.076} &\scalebox{1}{0.171} &\scalebox{1}{0.031} &\scalebox{1}{0.056}  &\scalebox{1}{0.055} &\scalebox{1}{0.117} &\scalebox{1}{0.049} &\scalebox{1}{0.112} \\
				\bottomrule
			\end{tabular}
		\end{small}
	\end{threeparttable}
\end{table*}

\subsection{Baselines.} 
We choose the latest and advanced time series model as our basic baselines, including Transformer-based models: OFA \cite{zhou2023one}, PatchTST \cite{nie2022time}, FEDFormer \cite{zhou2022fedformer}, CrossFormer \cite{Zhang2023CrossformerTU}, ETSFormer \cite{woo2022etsformer}; MLP-based models: Dlinear \cite{Zeng2022AreTE}, LightTS \cite{Zhang2022LessIM}, FreTS \cite{Yi2023FrequencydomainMA}; and convolution-based models: TimesNet \cite{wu2023timesnet}, SCINet \cite{Liu2021SCINetTS}, MICN \cite{Wang2023MICNML}. Besides, N-HiTS \cite{challu2022n} and N-BEATS \cite{oreshkin2019n} are used for short-term forecasting. Anomaly Transformer \cite{xu2021anomaly} is used for anomaly detection. XGBoost \cite{Chen2016XGBoostAS}, Rocket \cite{Dempster2020ROCKETEF}, LSTNet \cite{2018Modeling}, S4 \cite{gu2022efficiently} are used for classification.

Besides, we implement a strong baseline for the baseline operator in Equation (\ref{equ:baseline}), which equipped with the channel independent setting that converts the multivariate data into univariate data and patch design from \cite{nie2022time} to avoid under-estimating the results of baseline which have achieved comparable performance to the state-of-the-art methods as described below.

\subsection{Main Results.} 
Our proposed method achieves consistent state-of-the-art performance on five mainstream tasks. We provide the detailed results of each task below.

\paragraph{Long- and Short-term Forecasting.}
We adopt two types of benchmarks, including long-term and short-term forecasting, to evaluate the forecasting performance. For the long-term setting, we follow the benchmarks used in OFA \cite{zhou2023one} and PatchTST \cite{nie2022time}, including ETT \citep{haoyietal-informer-2021}, Electricity \citep{ecldata}, Traffic \citep{trafficdata}, Weather \citep{weatherdata}, and ILI \citep{ilidata}, covering five real-world applications. For the short-term setting, we adopt the M4 \citep{M4team2018dataset} dataset, which contains the yearly, quarterly, monthly collected univariate marketing data and so on. As shown in Table \ref{tab:long_term_forecasting_results},$\,$ \ref{tab:short_term_forecasting_results}, our ODFL shows great performance in both long- and short-term settings.

\begin{figure}[h]
	\begin{center}
	\includegraphics[width=1.025\linewidth]{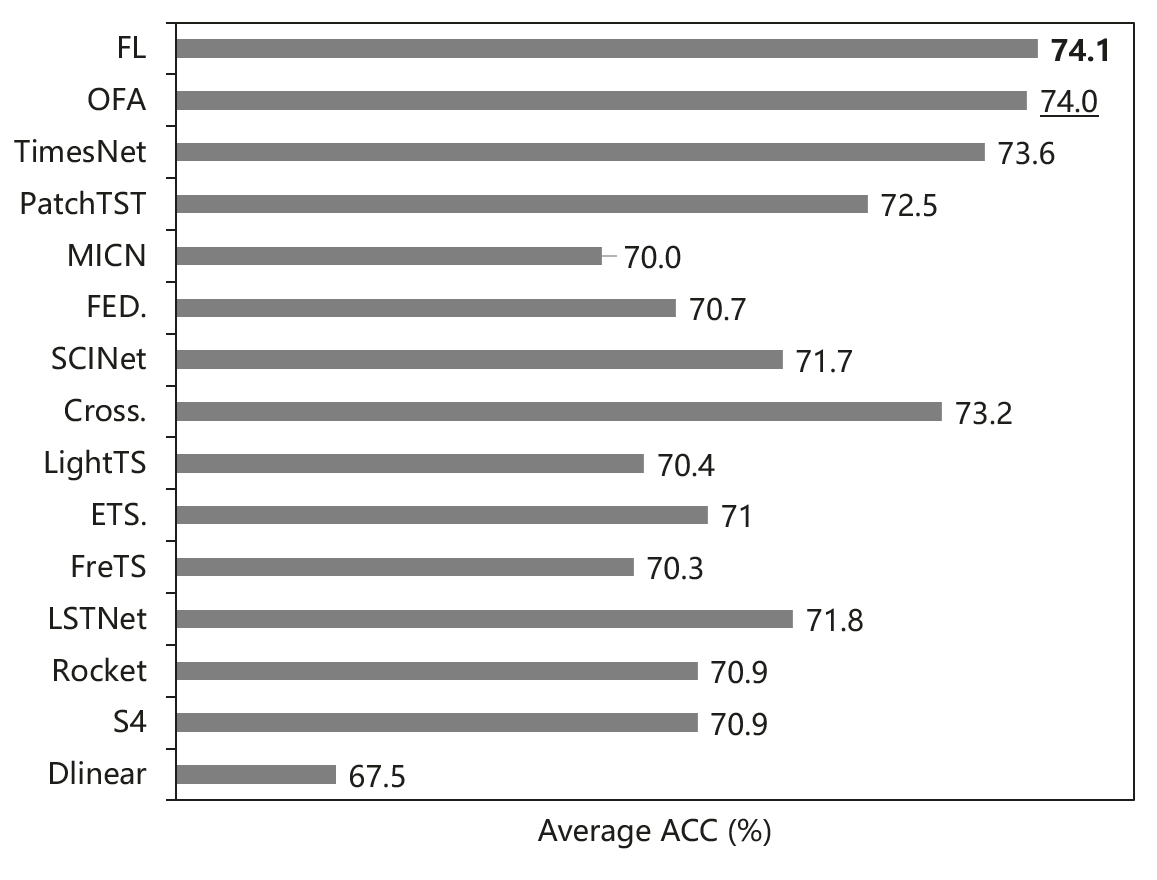}
	\end{center}
	\vspace{-8pt}
	\caption{Model comparison in classification. The results are averaged from 10 subsets of UEA. Higher accuracy indicates better performance. See Table \ref{tab:full_classification_results} in Appendix \ref{appendix:full} for full results.}
	\label{fig_classi}
\end{figure}

\paragraph{Imputation.}
We use the datasets from the electricity and weather scenarios as our benchmarks, including ETT \citep{haoyietal-informer-2021}, Electricity \citep{ecldata} and Weather \citep{weatherdata}, where the data-missing problem happens commonly. To compare the model capacity under different proportions of missing data, we randomly mask the time points in the ratio of $\{12.5\%, 25\%, 37.5\%, 50\%\}$. As shown in Table \ref{tab:imputation_results}, our ODFL surpasses the prior SOTA on all benchmarks with a large margin.

\begin{figure}[h]
	\begin{center}
		\includegraphics[width=1.025\linewidth]{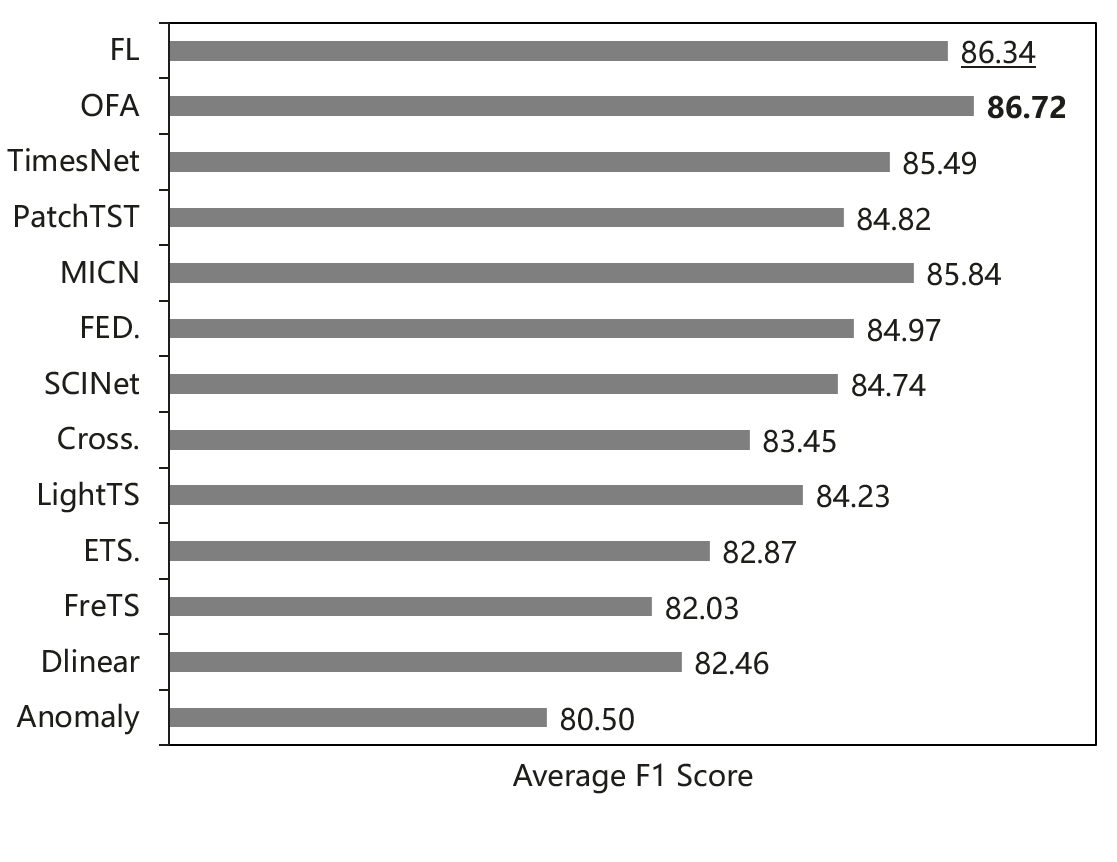}
	\end{center}
	\vspace{-16pt}
	\caption{Model comparison in anomaly detection tasks. The results are averaged from several datasets. Higher F1 score indicates better performance. See Table \ref{tab:full_anomaly_results} in Appendix \ref{appendix:full} for full results.}
	\label{fig_ano}
\end{figure}

\paragraph{Classification.}
Follow the benchmark proposed by TimesNet \cite{wu2023timesnet}, we use 10 multivariate sequence-level classification datasets from UEA Time Series Classification Archive \citep{Bagnall2018TheUM}, covering the gesture, action and audio recognition, medical diagnosis by heartbeat monitoring and other practical tasks to verify the model performance on high-level tasks. As shown in Figure \ref{fig_classi}, our ODFL achieves SOTA with an average accuracy of $74.1\%$.

\paragraph{Anomaly Detection.}
We compare models on five widely-used anomaly detection benchmarks: SMD \citep{Su2019RobustAD}, MSL \citep{Hundman2018DetectingSA}, SMAP \citep{Hundman2018DetectingSA}, SWaT \citep{DBLP:conf/cpsweek/MathurT16}, PSM \citep{DBLP:conf/kdd/AbdulaalLL21}, covering service monitoring, space \& earth exploration, and water treatment applications. Following the pre-processing methods in Anomaly Transformer \cite{xu2021anomaly}, we split the dataset into consecutive non-overlapping segments by sliding window. As shown in Figure \ref{fig_ano}, our ODFL also achieves SOTA performance, this demonstrates that our model satisfies the property that each task requires with high generalization ability.

\begin{table*}[htbp]
	\caption{Ablation of different modules evaluated on long-term forecasting task. The symbol $+$ means indicates that we add the corresponding element upon the upper line.}
	\label{tab:ablation}
	\centering
	\begin{threeparttable}
		\renewcommand{\multirowsetup}{\centering}
		\setlength{\tabcolsep}{2pt}
		\begin{tabular}{c|cc|cc|cc|cc|cc|cc|cc|cc}
			\toprule
			\multicolumn{1}{c}{\scalebox{1}{Dataset}} & 
			\multicolumn{2}{c}{\rotatebox{0}{\scalebox{1}{ETTm1}}} &
			\multicolumn{2}{c}{\rotatebox{0}{\scalebox{1}{ETTm2}}} &
			\multicolumn{2}{c}{\rotatebox{0}{\scalebox{1}{ETTh1}}} &
			\multicolumn{2}{c}{\rotatebox{0}{\scalebox{1}{ETTh2}}} &
			\multicolumn{2}{c}{\rotatebox{0}{\scalebox{1}{Electricity}}} &
			\multicolumn{2}{c}{\rotatebox{0}{\scalebox{1}{Traffic}}} & \multicolumn{2}{c}{\rotatebox{0}{\scalebox{1}{Weather}}} &  \multicolumn{2}{c}{\rotatebox{0}{\scalebox{1}{ILI}}} \\
			\cmidrule(lr){0-0} \cmidrule(lr){2-3} \cmidrule(lr){4-5}\cmidrule(lr){6-7} \cmidrule(lr){8-9}\cmidrule(lr){10-11}\cmidrule(lr){12-13}\cmidrule(lr){14-15}\cmidrule(lr){16-17}
			\multicolumn{1}{c}{\scalebox{1}{Metric}} & \scalebox{1}{MSE} & \scalebox{1}{MAE} & \scalebox{1}{MSE} & \scalebox{1}{MAE} & \scalebox{1}{MSE} & \scalebox{1}{MAE} & \scalebox{1}{MSE} & \scalebox{1}{MAE} & \scalebox{1}{MSE} & \scalebox{1}{MAE} & \scalebox{1}{MSE} & \scalebox{1}{MAE} & \scalebox{1}{MSE} & \scalebox{1}{MAE} & \scalebox{1}{MSE} & \scalebox{1}{MAE} \\
			\toprule
			\scalebox{1}{Baseline} &{\scalebox{1}{0.372}} &{\scalebox{1}{0.403}}  &\scalebox{1}{0.274} &\scalebox{1}{0.336} & \scalebox{1}{0.431} & \scalebox{1}{0.453} &\scalebox{1}{0.382} &\scalebox{1}{0.430} &\scalebox{1}{0.188} &\scalebox{1}{0.277} &\scalebox{1}{0.428} &\scalebox{1}{0.299} &\scalebox{1}{0.235} &\scalebox{1}{0.270} &\scalebox{1}{1.927} &\scalebox{1}{0.913}\\
			\midrule
			\scalebox{1}{\textit{$+$Partial.}} &\scalebox{1}{0.360} &\scalebox{1}{0.393} &\scalebox{1}{0.259} &\scalebox{1}{0.328} & \scalebox{1}{0.424} & \scalebox{1}{0.445} &\scalebox{1}{0.356} &\scalebox{1}{0.401} &\scalebox{1}{0.179} &\scalebox{1}{0.270} &\scalebox{1}{0.413} &\scalebox{1}{0.283} &\scalebox{1}{0.231} &\scalebox{1}{0.277} &\scalebox{1}{1.769} &\scalebox{1}{0.914}\\
			\midrule
			\scalebox{1}{\textit{$+$Salient.}} &{\scalebox{1}{0.354}} &{\scalebox{1}{0.389}} &{\scalebox{1}{0.255}} &{\scalebox{1}{0.320}} & \scalebox{1}{0.421} & \scalebox{1}{0.440} &\scalebox{1}{0.343} &\scalebox{1}{0.391} &\scalebox{1}{0.170} &\scalebox{1}{0.263} &\scalebox{1}{0.404} &\scalebox{1}{0.277} &\scalebox{1}{0.225} &\scalebox{1}{0.265} &\scalebox{1}{1.636} &\scalebox{1}{0.853}\\
			\midrule
			\scalebox{1}{\textit{$+$Adaptive.}}&\textbf{\scalebox{1}{0.343}} &\textbf{\scalebox{1}{0.378}}  &\textbf{\scalebox{1}{0.246}} &\textbf{\scalebox{1}{0.313}} &\textbf{\scalebox{1}{0.413}} 
			&\textbf{\scalebox{1}{0.432}} &\textbf{\scalebox{1}{0.331}} &\textbf{\scalebox{1}{0.384}} &\textbf{\scalebox{1}{0.158}} &\textbf{\scalebox{1}{0.253}} &\textbf{\scalebox{1}{0.388}} &\textbf{\scalebox{1}{0.265}} &\textbf{\scalebox{1}{0.221}} &\textbf{\scalebox{1}{0.260}} &\textbf{\scalebox{1}{1.431}} &\textbf{\scalebox{1}{0.792}} \\
			\bottomrule
		\end{tabular}
	\end{threeparttable}
\end{table*}

\begin{table*}[htbp]
	\caption{Ablation of the partial ratio on long-term forecasting task. A higher ratio means more channels are extracted.}
	\label{tab:abla_partial}
	\centering
	\begin{threeparttable}
		\renewcommand{\multirowsetup}{\centering}
		\setlength{\tabcolsep}{2pt}
		\begin{tabular}{c|cc|cc|cc|cc|cc|cc|cc|cc}
			\toprule
			\multicolumn{1}{c}{\scalebox{1}{Dataset}} & 
			\multicolumn{2}{c}{\rotatebox{0}{\scalebox{1}{ETTm1}}} &
			\multicolumn{2}{c}{\rotatebox{0}{\scalebox{1}{ETTm2}}} &
			\multicolumn{2}{c}{\rotatebox{0}{\scalebox{1}{ETTh1}}} &
			\multicolumn{2}{c}{\rotatebox{0}{\scalebox{1}{ETTh2}}} &
			\multicolumn{2}{c}{\rotatebox{0}{\scalebox{1}{Electricity}}} &
			\multicolumn{2}{c}{\rotatebox{0}{\scalebox{1}{Traffic}}} & \multicolumn{2}{c}{\rotatebox{0}{\scalebox{1}{Weather}}} &  \multicolumn{2}{c}{\rotatebox{0}{\scalebox{1}{ILI}}}  \\
			\cmidrule(lr){0-0} \cmidrule(lr){2-3} \cmidrule(lr){4-5}\cmidrule(lr){6-7} \cmidrule(lr){8-9}\cmidrule(lr){10-11}\cmidrule(lr){12-13}\cmidrule(lr){14-15}\cmidrule(lr){16-17}
			\multicolumn{1}{c}{\scalebox{1.2}{$r_{p}$}} & \scalebox{1}{MSE} & \scalebox{1}{MAE} & \scalebox{1}{MSE} & \scalebox{1}{MAE} & \scalebox{1}{MSE} & \scalebox{1}{MAE} & \scalebox{1}{MSE} & \scalebox{1}{MAE} & \scalebox{1}{MSE} & \scalebox{1}{MAE} & \scalebox{1}{MSE} & \scalebox{1}{MAE} & \scalebox{1}{MSE} & \scalebox{1}{MAE} & \scalebox{1}{MSE} & \scalebox{1}{MAE} \\
			\toprule
			\scalebox{1}{$100\%$} 
			&{\scalebox{1}{0.355}} &{\scalebox{1}{0.391}}  &\scalebox{1}{0.258} &\scalebox{1}{0.322} & \scalebox{1}{0.420} & \scalebox{1}{0.437} &\scalebox{1}{0.346} &\scalebox{1}{0.393} &\scalebox{1}{0.163} &\scalebox{1}{0.258} &\scalebox{1}{0.396} &\scalebox{1}{0.269} &\scalebox{1}{0.227} &\scalebox{1}{0.266} &\scalebox{1}{1.625} &\scalebox{1}{0.843} \\
			\midrule
			\scalebox{1}{$75\%$} 
			&\scalebox{1}{0.349} &\scalebox{1}{0.385} &\scalebox{1}{0.250} &\scalebox{1}{0.314} & \underline{\scalebox{1}{0.415}} &\textbf{\scalebox{1}{0.431}} &\underline{\scalebox{1}{0.335}} &\textbf{\scalebox{1}{0.383}} &\scalebox{1}{0.159} &\scalebox{1}{0.255} &\scalebox{1}{0.391} &\underline{\scalebox{1}{0.267}} &\scalebox{1}{0.222} &\scalebox{1}{0.264} &\scalebox{1}{1.547} &\scalebox{1}{0.819} \\
			\midrule
			\scalebox{1}{$50\%$} &\textbf{\scalebox{1}{0.343}} &\textbf{\scalebox{1}{0.378}} &\textbf{\scalebox{1}{0.246}} &\textbf{\scalebox{1}{0.313}} & \textbf{\scalebox{1}{0.413}} &\underline{\scalebox{1}{0.432}} &\textbf{\scalebox{1}{0.331}} &\underline{\scalebox{1}{0.384}} &\underline{\scalebox{1}{0.158}} &\underline{\scalebox{1}{0.253}} &\textbf{\scalebox{1}{0.388}} &\textbf{\scalebox{1}{0.265}} &\textbf{\scalebox{1}{0.221}} &\textbf{\scalebox{1}{0.260}} &\textbf{\scalebox{1}{1.431}} &\underline{\scalebox{1}{0.792}} \\
			\midrule
			\scalebox{1}{$25\%$}
			&\underline{\scalebox{1}{0.346}} &\underline{\scalebox{1}{0.379}}  &\underline{\scalebox{1}{0.247}} &\textbf{\scalebox{1}{0.313}} &{\scalebox{1}{0.416}} 
			&\scalebox{1}{0.435} &{\scalebox{1}{0.338}} &\scalebox{1}{0.388} &\textbf{\scalebox{1}{0.156}} &\textbf{\scalebox{1}{0.252}} &\underline{\scalebox{1}{0.389}} &\underline{\scalebox{1}{0.267}} &\textbf{\scalebox{1}{0.221}} &\underline{\scalebox{1}{0.263}} &\underline{\scalebox{1}{1.511}} &\textbf{\scalebox{1}{0.813}} \\
			\bottomrule
		\end{tabular}
	\end{threeparttable}
\end{table*}

\subsection{Ablation Study}
\label{sec:abla}
\paragraph{Ablation of the Block Design.} To validate the effectiveness of our designation within the proposed ODFL block, we ablate the important elements of our design on long-term forecasting task. Results are shown on Table \ref{tab:ablation}. 
As we can see in the first line, the baseline with the global filter described above has inferior performance compared with self attention in the time domain \cite{nie2022time} similar to the phenomena observed in \cite{Zhou2022FiLMFI}. 
After adding our partial operation on the channel dimension, the model can outperform the baseline model with fewer parameters as shown in the second line. 
To better incorporate with time series properties, we introduce the sparsity operation on the salient frequency parts which improves the model performance by extracting the most informative elements while avoiding overfitting to the low SNR parts as shown in the third line.
Finally, thanks to our semantic-aware adaptive filter, the performance increases consistently by handing the varied semantics as shown in the last line which represents our proposed ODFL model.

\paragraph{Varying Partial Ratio.} To verify the impact of the partial ratio $r_{p}$ in Equation (\ref{equ:partial}), we conduct ablation experiments on long-term forecasting task in Table \ref{tab:abla_partial}, which empirically shows that the partial operation can improve performance when vary in a wide range consistently. The result verifies the previous statement.
As a result, for the partial ratio $r_{p}$, we set it to $50\%$ for all variants by default for both effectiveness and efficiency.

\paragraph{Varying Sparsity Ratio.} We conduct a brief ablation study on the value of sparsity ratio $r_{s}$ in Equation (\ref{equ:salient}), results are summarized in Table \ref{tab:abla_sparsity}. A too large sparsity ratio $r_{s}$ would make the kernel degrade to a regular global filter which will affected by the noisy frequency bands, while a too small value would render our filter less effective in capturing the spectrum features. Thus, we simply set $r_{s}$ to $75\%$ by default for simplicity. Besides, we compare other frequency modes selection strategy in Appendix \ref{abla:selection_strategy}.

\begin{table*}[htbp]
	\caption{Ablation of the sparsity ratio on long-term forecasting task. A higher ratio means more frequency bands are filtered or extracted.}
	\label{tab:abla_sparsity}
	\centering
	\begin{threeparttable}
		\renewcommand{\multirowsetup}{\centering}
		\setlength{\tabcolsep}{2pt}
		\begin{tabular}{c|cc|cc|cc|cc|cc|cc|cc|cc}
			\toprule
			\multicolumn{1}{c}{\scalebox{1}{Dataset}} & 
			\multicolumn{2}{c}{\rotatebox{0}{\scalebox{1}{ETTm1}}} &
			\multicolumn{2}{c}{\rotatebox{0}{\scalebox{1}{ETTm2}}} &
			\multicolumn{2}{c}{\rotatebox{0}{\scalebox{1}{ETTh1}}} &
			\multicolumn{2}{c}{\rotatebox{0}{\scalebox{1}{ETTh2}}} &
			\multicolumn{2}{c}{\rotatebox{0}{\scalebox{1}{Electricity}}} &
			\multicolumn{2}{c}{\rotatebox{0}{\scalebox{1}{Traffic}}} & \multicolumn{2}{c}{\rotatebox{0}{\scalebox{1}{Weather}}} & \multicolumn{2}{c}{\rotatebox{0}{\scalebox{1}{ILI}}}  \\
			\cmidrule(lr){0-0} \cmidrule(lr){2-3} \cmidrule(lr){4-5}\cmidrule(lr){6-7} \cmidrule(lr){8-9}\cmidrule(lr){10-11}\cmidrule(lr){12-13}\cmidrule(lr){14-15}\cmidrule(lr){16-17}
			\multicolumn{1}{c}{\scalebox{1.2}{$r_{s}$}} & \scalebox{1}{MSE} & \scalebox{1}{MAE} & \scalebox{1}{MSE} & \scalebox{1}{MAE} & \scalebox{1}{MSE} & \scalebox{1}{MAE} & \scalebox{1}{MSE} & \scalebox{1}{MAE} & \scalebox{1}{MSE} & \scalebox{1}{MAE} & \scalebox{1}{MSE} & \scalebox{1}{MAE} & \scalebox{1}{MSE} & \scalebox{1}{MAE} & \scalebox{1}{MSE} & \scalebox{1}{MAE} \\
			\toprule
			\scalebox{1}{$100\%$} 
			&{\scalebox{1}{0.350}} &{\scalebox{1}{0.384}}  &\scalebox{1}{0.251} &\scalebox{1}{0.317} & \underline{\scalebox{1}{0.414}} & \underline{\scalebox{1}{0.435}} &\scalebox{1}{0.341} &\scalebox{1}{0.390} &\scalebox{1}{0.166} &\scalebox{1}{0.261} &\scalebox{1}{0.395} &\scalebox{1}{0.268} &\scalebox{1}{0.225} &\scalebox{1}{0.263} &\scalebox{1}{1.501} &\scalebox{1}{0.808} \\
			\midrule
			\scalebox{1}{$75\%$} 
			&\textbf{\scalebox{1}{0.343}} &\textbf{\scalebox{1}{0.378}} &\textbf{\scalebox{1}{0.246}} &\textbf{\scalebox{1}{0.313}} 
			&\textbf{\scalebox{1}{0.413}} 
			&\textbf{\scalebox{1}{0.432}} &\textbf{\scalebox{1}{0.331}} &\textbf{\scalebox{1}{0.384}} &\textbf{\scalebox{1}{0.158}} &\textbf{\scalebox{1}{0.253}} &\textbf{\scalebox{1}{0.388}} &\textbf{\scalebox{1}{0.265}} &\textbf{\scalebox{1}{0.221}} &\textbf{\scalebox{1}{0.260}} &\textbf{\scalebox{1}{1.431}} &\textbf{\scalebox{1}{0.792}}\\
			\midrule
			\scalebox{1}{$50\%$}
			&\underline{\scalebox{1}{0.344}} &\underline{\scalebox{1}{0.380}}  &\underline{\scalebox{1}{0.250}} &\underline{\scalebox{1}{0.314}} &\scalebox{1}{0.415} & \scalebox{1}{0.436} &\underline{\scalebox{1}{0.333}} &\underline{\scalebox{1}{0.387}} &\underline{\scalebox{1}{0.161}} &\underline{\scalebox{1}{0.255}} &\underline{\scalebox{1}{0.389}} &\underline{\scalebox{1}{0.267}} &\underline{\scalebox{1}{0.223}} &\underline{\scalebox{1}{0.261}} &\underline{\scalebox{1}{1.463}} &\underline{\scalebox{1}{0.795}} \\
			\bottomrule
		\end{tabular}
	\end{threeparttable}
\end{table*}

\begin{table*}[htbp]
	\caption{Comparison results of SGConv module. The symbol $+$ means indicates that we add the corresponding element upon the baseline.}
	\label{tab:sgconv}
	\centering
	\begin{threeparttable}
			\renewcommand{\multirowsetup}{\centering}
			\setlength{\tabcolsep}{2pt}
			\begin{tabular}{c|cc|cc|cc|cc|cc|cc|cc|cc}
				\toprule
				\multicolumn{1}{c}{\scalebox{1}{Dataset}} & 
				\multicolumn{2}{c}{\rotatebox{0}{\scalebox{1}{ETTm1}}} &
				\multicolumn{2}{c}{\rotatebox{0}{\scalebox{1}{ETTm2}}} &
				\multicolumn{2}{c}{\rotatebox{0}{\scalebox{1}{ETTh1}}} &
				\multicolumn{2}{c}{\rotatebox{0}{\scalebox{1}{ETTh2}}} &
				\multicolumn{2}{c}{\rotatebox{0}{\scalebox{1}{Electricity}}} &
				\multicolumn{2}{c}{\rotatebox{0}{\scalebox{1}{Traffic}}} & \multicolumn{2}{c}{\rotatebox{0}{\scalebox{1}{Weather}}} &  \multicolumn{2}{c}{\rotatebox{0}{\scalebox{1}{ILI}}} \\
				\cmidrule(lr){0-0} \cmidrule(lr){2-3} \cmidrule(lr){4-5}\cmidrule(lr){6-7} \cmidrule(lr){8-9}\cmidrule(lr){10-11}\cmidrule(lr){12-13}\cmidrule(lr){14-15}\cmidrule(lr){16-17}
				\multicolumn{1}{c}{\scalebox{1}{Metric}} & \scalebox{1}{MSE} & \scalebox{1}{MAE} & \scalebox{1}{MSE} & \scalebox{1}{MAE} & \scalebox{1}{MSE} & \scalebox{1}{MAE} & \scalebox{1}{MSE} & \scalebox{1}{MAE} & \scalebox{1}{MSE} & \scalebox{1}{MAE} & \scalebox{1}{MSE} & \scalebox{1}{MAE} & \scalebox{1}{MSE} & \scalebox{1}{MAE} & \scalebox{1}{MSE} & \scalebox{1}{MAE} \\
				\toprule
				\scalebox{1}{Baseline} &{\scalebox{1}{0.372}} &{\scalebox{1}{0.403}}  &\scalebox{1}{0.274} &\scalebox{1}{0.336} & \underline{\scalebox{1}{0.431}} & \scalebox{1}{0.453} &\scalebox{1}{0.382} &\underline{\scalebox{1}{0.430}} &\scalebox{1}{0.188} &\scalebox{1}{0.277} &\underline{\scalebox{1}{0.428}} &\underline{\scalebox{1}{0.299}} &\scalebox{1}{0.235} &\underline{\scalebox{1}{0.270}} &\scalebox{1}{1.927} &{\scalebox{1}{0.913}}\\
				\midrule
				\scalebox{1}{\textit{$+$SGConv.}} &\underline{\scalebox{1}{0.368}} &\underline{\scalebox{1}{0.400}} &\underline{\scalebox{1}{0.269}} &\underline{\scalebox{1}{0.334}} & {\scalebox{1}{0.432}} &\underline{\scalebox{1}{0.450}} &\underline{\scalebox{1}{0.380}} &\scalebox{1}{0.433} &\underline{\scalebox{1}{0.180}} &\underline{\scalebox{1}{0.276}} &{\scalebox{1}{0.430}} &\scalebox{1}{0.301} &\underline{\scalebox{1}{0.230}} &\underline{\scalebox{1}{0.270}} &\underline{\scalebox{1}{1.905}} &\underline{\scalebox{1}{0.903}} \\
				\midrule
				\scalebox{1}{ODFL}&\textbf{\scalebox{1}{0.343}} &\textbf{\scalebox{1}{0.378}}  &\textbf{\scalebox{1}{0.246}} &\textbf{\scalebox{1}{0.313}} &\textbf{\scalebox{1}{0.413}} 
				&\textbf{\scalebox{1}{0.432}} &\textbf{\scalebox{1}{0.331}} &\textbf{\scalebox{1}{0.384}} &\textbf{\scalebox{1}{0.158}} &\textbf{\scalebox{1}{0.253}} &\textbf{\scalebox{1}{0.388}} &\textbf{\scalebox{1}{0.265}} &\textbf{\scalebox{1}{0.221}} &\textbf{\scalebox{1}{0.260}} &\textbf{\scalebox{1}{1.431}} &\textbf{\scalebox{1}{0.792}} \\
				\bottomrule
			\end{tabular}
	\end{threeparttable}
\end{table*}

\subsection{Model Analysis}
\label{sec:analysis}

\paragraph{Compared with SGConv Module.} SGConv \cite{Li2022WhatMC} propose two basic principles for the global kernel network S4 \cite{gu2022efficiently} with a special parameterization based on
the Cauchy kernel: the number of parameters should scale sub-linearly with sequence length, and the kernel needs to satisfy a decaying structure. 
This brings performance improvement on the Long Range Arena (LRA) benchmark \cite{Tay2020LongRA}.
However, this boost is not much on time series tasks as shown in Table \ref{tab:sgconv}.

Since simply transferring S4 to time series forecasting task cause much weaker performance than SoTA as shown in \cite{Zhou2022FiLMFI}, we compare SGConv module under our strong baseline for fair comparison.
As shown in the second line on Table \ref{tab:sgconv}, replacing the global kernel with SGConv kernel follow the official implementation\footnote{https://github.com/ctlllll/SGConv.} brings minimal improvement.
This is mainly because the LRA task needs more inductive biases for modeling its much long sequence ($1K$ to $16K$) and complex data structure. For example, all existing models failed on modeling the spatial-level sequence modeling task Pathfinder-X in the LRA benchmark except S4 which requires sophisticated parameterization and initialization schemes that combine the wisdom from several prior works. Also, the exponential decay property of the spectrum of matrix powers for time series signal is much easier to learn from our semantic-aware kernel and un-salient frequency selection strategy as visualized in Appendix \ref{appen:vis_kernel}. Moreover, we find that different from other field, the initialization of the kernel shows minimal impact for time series task as detailed in Appendix \ref{appen:init}.

\paragraph{Compared with TimesNet.} TimesNet \cite{wu2023timesnet} also demonstrates excellent generality in time series analysis tasks as well as taking advantage of frequency representation by capturing unsalient period-variation. However, they accomplish this goal by transforming the 1D time series into 2D space and model the cyclic information by 2D convolution operation. This additional transformation and aggregation operation makes TimesNet heavy-weight and hard to train. Moreover, it can model limited significant frequencies only and will neglect valuable information within the frequency domain. On the contrary, as described in Appendix \ref{proof:seasonality}, our proposed ODFL can model every significant frequency simply by multiplication operation.

\paragraph{Benefiting from Representation Learning.} 
Sophisticated deep models with larger capacity than linear \cite{Zeng2022AreTE} or classical statistical \cite{2008Forecasting} models that capture abstract representation of the data can benefiting from self-supervised learning techniques \cite{nie2022time,Eldele2021TimeSeriesRL,Yang2022UnsupervisedTR,Zhang2022SelfSupervisedCP}. In this part, we follow the masked self-supervised learning method \cite{He2021MaskedAA} and setting that we randomly remove a portion of input sequence and then the model is trained to recover the missing contents from PatchTST \cite{nie2022time} on all the long-term forecasting training datasets. After that we fine-tune the pre-trained model on the corresponding dataset. The result in Table \ref{tab:self} show that ODFL can be further enhanced by representation learning.

\begin{table}[htbp]
	\centering
	\caption{The results of representation learning.}
	\label{tab:self}
	\begin{tabular}{cc|c|cc|cc}
		\toprule
		&\multicolumn{2}{c|}{Models} & \multicolumn{2}{c|}{Fine-tuning} & \multicolumn{2}{c}{Supervised}\\
		\midrule
		&\multicolumn{2}{c|}{Metric}&MSE&MAE&MSE&MAE\\
		\midrule
		&\multirow{4}*{\rotatebox{90}{ETTm1}}& 96 & \textbf{0.280} & \textbf{0.341} & 0.282 & 0.343 \\
		&\multicolumn{1}{c|}{}& 192 & \textbf{0.319} & \textbf{0.365} & 0.322 & 0.368 \\
		&\multicolumn{1}{c|}{}& 336 & \textbf{0.355} & \textbf{0.385} & 0.356 & 0.388 \\
		&\multicolumn{1}{c|}{}& 720 & \textbf{0.405} & \textbf{0.401} & 0.412 & 0.411 \\
		\midrule
		&\multirow{4}*{\rotatebox{90}{Weather}}& 96 & \textbf{0.140} & \textbf{0.195} & 0.145 & 0.198 \\
		&\multicolumn{1}{c|}{}& 192 & \textbf{0.184} & \textbf{0.230} & 0.190 & 0.238 \\
		&\multicolumn{1}{c|}{}& 336 & \textbf{0.236} & \textbf{0.275} & 0.240 & 0.279 \\
		&\multicolumn{1}{c|}{}& 720 & \textbf{0.300} & \textbf{0.318} & 0.309 & 0.330 \\
		\midrule
		&\multirow{4}*{\rotatebox{90}{Electricity}}& 96 & \textbf{0.126} & \textbf{0.221} & 0.128 & 0.224 \\
		&\multicolumn{1}{c|}{}& 192 & \textbf{0.144} & \textbf{0.240} & 0.145 & \textbf{0.240} \\
		&\multicolumn{1}{c|}{}& 336 & \textbf{0.160} & \textbf{0.256} & 0.162 & 0.260 \\
		&\multicolumn{1}{c|}{}& 720 & \textbf{0.190} & \textbf{0.281} & 0.195 & 0.288 \\
		\midrule
		&\multirow{4}*{\rotatebox{90}{Traffic}}& 96 & \textbf{0.358} & \textbf{0.390} & 0.369 & 0.402 \\
		&\multicolumn{1}{c|}{}& 192 & \textbf{0.401} & \textbf{0.412} & 0.409 & 0.425 \\
		&\multicolumn{1}{c|}{}& 336 & \textbf{0.418} & \textbf{0.430} & 0.425 & 0.441 \\
		&\multicolumn{1}{c|}{}& 720 & \textbf{0.442} & \textbf{0.451} & 0.449 & 0.461 \\
		\bottomrule
	\end{tabular}
\end{table}

\section{Conclusion}
In this work, we proposes the Frequency Learner as a task-general model for time series analysis. We propose three key feature properties for modeling time series feature in the frequency domain: channel redundancy, sparse frequency distribution, and semantic diversity. By utilizing our semantic-aware global filter with sparse operation along both the channel and frequency dimension, ODFL achieves state-of-the-art performance on five mainstream tasks.

In the future work, we will further explore the large-scale self-supervised pre-training methods upon our proposed ODFL model to achieve better task-general ability.

\nocite{langley00}

\bibliography{example_paper}
\bibliographystyle{icml2024}

\newpage
\appendix
\onecolumn

\section{Setting}
\label{appen:setting}
\subsection{Benchmark}
\label{appen:datasets}
We provide the detailed dataset descriptions in Table \ref{tab:dataset}. 

We adopt the mean square error (MSE) and mean absolute error (MAE) for long-term forecasting and imputations. For anomaly detection, we adopt the F1-score as the metric. For the short-term forecasting, we adopt the symmetric mean absolute percentage error (SMAPE), mean absolute scaled error (MASE) and overall weighted average (OWA) as the metrics:

\begin{align} 
	\label{equ:metrics}
	\text{SMAPE} &= \frac{200}{H} \sum_{i=1}^H \frac{|\mathbf{X}_{i} - \widehat{\mathbf{X}}_{i}|}{|\mathbf{X}_{i}| + |\widehat{\mathbf{X}}_{i}|},\\
	\text{MAPE} &= \frac{100}{H} \sum_{i=1}^H \frac{|\mathbf{X}_{i} - \widehat{\mathbf{X}}_{i}|}{|\mathbf{X}_{i}|}, \\
	\text{MASE} &= \frac{1}{H} \sum_{i=1}^H \frac{|\mathbf{X}_{i} - \widehat{\mathbf{X}}_{i}|}{\frac{1}{H-m}\sum_{j=m+1}^{H}|\mathbf{X}_j - \mathbf{X}_{j-m}|},\\
	\text{OWA} &= \frac{1}{2} \left[ \frac{\text{SMAPE}}{\text{SMAPE}_{\textrm{Naïve2}}}  + \frac{\text{MASE}}{\text{MASE}_{\textrm{Naïve2}}}  \right].
\end{align}

where $m$ is the periodicity of the data. $\mathbf{X},\widehat{\mathbf{X}}\in\mathbb{R}^{H\times C}$ are the ground truth and prediction results of the future with $H$ time pints and $C$ dimensions. $\mathbf{X}_{i}$ means the $i$-th future time point. 

\begin{table*}[htbp]
	\vspace{-9pt}
	\caption{Dataset descriptions. The dataset size is organized in (Train, Validation, Test).}\label{tab:dataset}
	\centering
	\begin{threeparttable}
			\renewcommand{\multirowsetup}{\centering}
			\setlength{\tabcolsep}{4pt}
			\begin{tabular}{c|l|c|c|c|c}
				\toprule
				Tasks & Dataset & Dim & Series Length & Dataset Size & \scalebox{1}{Information (Frequency)} \\
				\toprule
				& ETTm1, ETTm2 & 7 & \scalebox{1}{\{96, 192, 336, 720\}} & (34465, 11521, 11521) & \scalebox{1}{Electricity (15 mins)}\\
				\cmidrule{2-6}
				& ETTh1, ETTh2 & 7 & \scalebox{1}{\{96, 192, 336, 720\}} & (8545, 2881, 2881) & \scalebox{1}{Electricity (15 mins)} \\
				\cmidrule{2-6}
				Long-term & Electricity & 321 & \scalebox{1}{\{96, 192, 336, 720\}} & (18317, 2633, 5261) & \scalebox{0.8}{Electricity (Hourly)} \\
				\cmidrule{2-6}
				Forecasting & Traffic & 862 & \scalebox{1}{\{96, 192, 336, 720\}} & (12185, 1757, 3509) & \scalebox{1}{Transportation (Hourly)} \\
				\cmidrule{2-6}
				& Weather & 21 & \scalebox{1}{\{96, 192, 336, 720\}} & (36792, 5271, 10540) & \scalebox{1}{Weather (10 mins)} \\
				\cmidrule{2-6}
				& ILI & 7 & \scalebox{1}{\{24, 36, 48, 60\}} & (617, 74, 170) & \scalebox{1}{Illness (Weekly)} \\
				\midrule
				& M4-Yearly & 1 & 6 & (23000, 0, 23000) & \scalebox{1}{Demographic} \\
				\cmidrule{2-6}
				& M4-Quarterly & 1 & 8 & (24000, 0, 24000) & \scalebox{1}{Finance} \\
				\cmidrule{2-6}
				Short-term & M4-Monthly & 1 & 18 & (48000, 0, 48000) & \scalebox{1}{Industry} \\
				\cmidrule{2-6}
				Forecasting & M4-Weakly & 1 & 13 & (359, 0, 359) & \scalebox{1}{Macro} \\
				\cmidrule{2-6}
				& M4-Daily & 1 & 14 & (4227, 0, 4227) & \scalebox{1}{Micro} \\
				\cmidrule{2-6}
				& M4-Hourly & 1 &48 & (414, 0, 414) & \scalebox{1}{Other} \\
				\midrule
				\multirow{5}{*}{Imputation} & ETTm1, ETTm2 & 7 & 96 & (34465, 11521, 11521) & \scalebox{1}{Electricity (15 mins)} \\
				\cmidrule{2-6}
				& ETTh1, ETTh2 & 7 & 96 & (8545, 2881, 2881) & \scalebox{1}{Electricity (15 mins)}\\
				\cmidrule{2-6}
				& Electricity & 321 & 96 & (18317, 2633, 5261) & \scalebox{1}{Electricity (15 mins)}\\
				\cmidrule{2-6}
				& Weather & 21 & 96 & (36792, 5271, 10540) & \scalebox{1}{Weather (10 mins)} \\
				\midrule
				& \scalebox{1}{EthanolConcentration} & 3 & 1751 & (261, 0, 263) & \scalebox{1}{Alcohol Industry}\\
				\cmidrule{2-6}
				& \scalebox{1}{FaceDetection} & 144 & 62 & (5890, 0, 3524) & \scalebox{1}{Face (250Hz)}\\
				\cmidrule{2-6}
				& \scalebox{1}{Handwriting} & 3 & 152 & (150, 0, 850) & \scalebox{1}{Handwriting}\\
				\cmidrule{2-6}
				& \scalebox{1}{Heartbeat} & 61 & 405 & (204, 0, 205)& \scalebox{1}{Heart Beat}\\
				\cmidrule{2-6}
				Classification & \scalebox{0.8}{JapaneseVowels} & 12 & 29 & (270, 0, 370) & \scalebox{1}{Voice}\\
				\cmidrule{2-6}
				(UEA) & \scalebox{1}{PEMS-SF} & 963 & 144 & (267, 0, 173) & \scalebox{1}{Transportation (Daily)}\\
				\cmidrule{2-6}
				& \scalebox{1}{SelfRegulationSCP1} & 6 & 896 & (268, 0, 293) & \scalebox{1}{Health (256Hz)}\\
				\cmidrule{2-6}
				& \scalebox{1}{SelfRegulationSCP2} & 7 & 1152 & (200, 0, 180) & \scalebox{1}{Health (256Hz)}\\
				\cmidrule{2-6}
				& \scalebox{1}{SpokenArabicDigits} & 13 & 93 & (6599, 0, 2199) & \scalebox{1}{Voice (11025Hz)}\\
				\cmidrule{2-6}
				& \scalebox{1}{UWaveGestureLibrary} & 3 & 315 & (120, 0, 320) & \scalebox{1}{Gesture}\\
				\midrule
				& SMD & 38 & 100 & (566724, 141681, 708420) & \scalebox{1}{Server Machine} \\
				\cmidrule{2-6}
				Anomaly & MSL & 55 & 100 & (44653, 11664, 73729) & \scalebox{1}{Spacecraft} \\
				\cmidrule{2-6}
				Detection & SMAP & 25 & 100 & (108146, 27037, 427617) & \scalebox{1}{Spacecraft} \\
				\cmidrule{2-6}
				& SWaT & 51 & 100 & (396000, 99000, 449919) & \scalebox{1}{Infrastructure} \\
				\cmidrule{2-6}
				& PSM & 25 & 100 & (105984, 26497, 87841)& \scalebox{1}{Server Machine} \\
				\bottomrule
			\end{tabular}
	\end{threeparttable}
\end{table*}

\begin{table*}[thbp]
	\caption{Experiment configuration of our ODFL.}\label{tab:model_config}
	\vspace{-1pt}
	\centering
	\begin{threeparttable}
			\renewcommand{\multirowsetup}{\centering}
			\setlength{\tabcolsep}{3pt}
			\begin{tabular}{c|c|c|c|c|c|c|c}
				\toprule
				\multirow{2}{*}{Tasks / Configurations} & \multicolumn{3}{c|}{Model Hyper-parameter} & \multicolumn{4}{c}{Training Process} \\
				\cmidrule(lr){2-4}\cmidrule(lr){5-8}
				& Layers & $d_{\text{min}}$ $^\dag$ & $d_{\text{max}}$ $^\dag$  & LR$^\ast$ & Loss & Batch Size & Epochs\\
				\toprule
				Long-term Forecasting & 2 & 32 & 512 & $10^{-4}$ & MSE & 32 & 100 \\
				\midrule
				Short-term Forecasting & 2 & 16 & 64 & $10^{-3}$ & SMAPE & 16 & 100 \\
				\midrule
				Imputation & 2 & 64 & 128 & $10^{-3}$ & MSE & 16 & 100 \\
				\midrule
				Classification & 2 & 32 & 64 & $10^{-3}$ & Cross Entropy & 16 & 30 \\
				\midrule
				Anomaly Detection & 3 & 32 & 128 & $10^{-4}$ & MSE & 128 & 10 \\
				\bottomrule
			\end{tabular}
			\begin{tablenotes}
				\footnotesize
				\item[] $\dag$ $d_{\text{model}}=\min\{\max\{2^{\lceil\log C\rceil}, d_{\text{min}}\},d_{\text{max}}\}$, where $C$ is input series dimension.
			\end{tablenotes}
	\end{threeparttable}
	\vspace{-7pt}
\end{table*}

\subsection{Implementation Details}
\label{appen:imp_details}
All experiments are implemented in PyTorch $1.10.0$\footnote{The index operation with complex dtype data for implementing Equation (\ref{equ:salient}) in the lower Pytorch version does not support automatic differentiation.} \cite{Paszke2019PyTorchAI} and conducted on a single NVIDIA V100 32GB GPU. All the experiments are repeated 3 times with different seeds and the means of the metrics are reported as the final results. Our implementation is bulit on the Time Series Library framework \cite{Time-Series-Library}. The linear head with a flatten layer is used to obtain the final prediction. We use patch length $P=16$ and stride $S=8$ for all experiments follow \cite{nie2022time} except that for short-term forecasting task with very short sequence we simply set patch length $P=1$ and stride $S=1$. Follow \cite{wu2023timesnet}, we select the $d_{\text{model}}$ based on the input series dimension $C$ by $\min\{\max\{2^{\lceil\log C\rceil}, d_{\text{min}}\},d_{\text{max}}\}$ to handle various dimensions of input sequence. The partial ratio is set to $50\%$ and the sparsity ratio is set to $75\%$. The FFN ratio is set to $4$. We use the Adam \cite{2014Adam} optimizer with $(\beta_1, \beta_2)$ as (0.9, 0.999). The initial learning rate is listed in Table \ref{tab:model_config}. Learning rate is dropped by half every epoch. An early stopping counter is employed to stop the training process after ten epochs if no loss degradation on the valid set is observed.

We collect some baseline results from OFA \cite{zhou2023one} and re-run other baselines under the same setting in \cite{wu2023timesnet} follow the official implementation, where all the baselines are re-run with various hyper-parameters and the best results are chosen to avoid under-estimating the baselines. 

Besides, for the long-term forecasting task we also re-run them with vary input length $L \in \{24, 48, 96, 192, 336, 512, 720\}$ follow \cite{zhou2023one,nie2022time} and choose the best results.
Moreover, we re-run FreTS \cite{Yi2023FrequencydomainMA} with ReVIN normalizaiton introduced by its official code\footnote{https://github.com/aikunyi/FreTS.} under the common data loader setting\footnote{https://github.com/yuqinie98/PatchTST and https://github.com/thuml/Time-Series-Library.} to have a fair comparison.

\section{Discussion}
\label{appen:discussion}
\subsection{The DFT of real signals are conjugate symmetric}
\label{proof:real_signals}
Given a real signal $\bm{x}\in \mathbb{R}^{n}$, the DFT of it $\bm{X}\in \mathbb{R}^{N}$ is conjugate symmetric, which can be proved as follows:
\begin{equation}
	\begin{aligned}
		\bm{X}[N - k] = \sum_{n=0}^{N-1}\bm{x}[n]e^{-j(2\pi / N)(N-k)n} \\
		=\sum_{n=0}^{N-1}\bm{x}[n]e^{j(2\pi / N)kn}=\bm{X}[k].
	\end{aligned}
\end{equation}

\subsection{The filter can serve as a global mixer}
\label{proof:circular_conv}
We firstly give the definition of circular convolution of a signal $\bm{x}[n]$ and a filter $\bm{k}[n]$ in the time domain:
\begin{equation}
	\label{equ:circular_conv}
	y[n] = \sum_{m=0}^{N-1}\bm{k}[m]\bm{x}[((n-m))_N],
\end{equation}

where $((n))_N$ denotes $n$ modulo $N$. Then multiplication in the frequency domain with a global kernel $\bm{K}[n]$ can be rewritten as follows:
\begin{equation}
	\begin{aligned}
		&\bm{K}[k]\bm{X}[k]\\
		&=\bm{K}[k]\sum_{n=0}^{N-1}\bm{x}[n]e^{-j(2\pi/N)kn}\\
		&=\bm{K}[k]\left(\sum_{n=0}^{N-m-1}\bm{x}[n]e^{-j(2\pi/N)kn} + \sum_{n=N-m}^{N-1}\bm{x}[n]e^{-j(2\pi/N)kn}\right)\\
		&=\bm{K}[k] \Bigg(
		\sum_{n=m}^{N-1}\bm{x}[n-m]e^{-j(2\pi/N)k(n-m)}\\
		&\qquad + \sum_{n=0}^{m-1}\bm{x}[n-m + N]e^{-j(2\pi/N)k(n-m)} \Bigg) \\
		&=\sum_{m=0}^{N-1}\bm{k}[m]e^{-j(2\pi/N)km}\sum_{n=0}^{N-1}\bm{x}[((n-m))_N]e^{-j(2\pi/N)k(n-m)} \\
		&=\sum_{n=0}^{N-1} \sum_{m=0}^{N-1}\bm{k}[m]\bm{x}[((n-m))_N]e^{-j(2\pi/N)kn} \\
		&=Y[k],
	\end{aligned}
\end{equation}

where $Y[k]$ is the DFT of $y[n]$. Thus, the operation of multiplying our sparse kernel with the salient frequencies $\left\{f_1, \cdots, f_k\right\}$ is equivalent to multiplying a global kernel $\hat{\bm{K}} \in\mathbb{R}^{k \times F}$ with the full frequency feature $\bm{X} \in\mathbb{R}^{F \times C}$
, where for frequencies $f \notin \left\{f_1, \cdots, f_k\right\}$, the kernel values are fixed to $1$. Thus, our large sparse kernel is also equivalent to a global depth-wise circular convolution in the time domain, which can model global trends.

\subsection{ODFL learns seasonality}
\label{proof:seasonality}
DFT represents the time series by a set of elementary functions called basis, where each basis $X[k]$ represents the spectrum of the sequence $x[n]$ at the frequency $\omega_k = 2\pi k/N$, which is corresponding to the period length $\lceil\frac{T}{j}\rceil$, as discussed in Equation (\ref{equ:dft}). Thus each element of our kernel $\bm{K}$ corresponds to a spectrum of the sequence. Benefiting from this representation, our method can model both long- and short-term seasonality explicitly \cite{1980The,2019Identifying}.

\begin{table*}[ht]
	\caption{Ablation study of the frequency selection strategies.}
	\label{tab:selecltion}
	\centering
	\begin{threeparttable}
			\renewcommand{\multirowsetup}{\centering}
			\setlength{\tabcolsep}{2pt}
			\begin{tabular}{c|cc|cc|cc|cc|cc|cc|cc|cc}
				\toprule
				\multicolumn{1}{c}{\scalebox{1}{Dataset}} & 
				\multicolumn{2}{c}{\rotatebox{0}{\scalebox{1}{ETTm1}}} &
				\multicolumn{2}{c}{\rotatebox{0}{\scalebox{1}{ETTm2}}} &
				\multicolumn{2}{c}{\rotatebox{0}{\scalebox{1}{ETTh1}}} &
				\multicolumn{2}{c}{\rotatebox{0}{\scalebox{1}{ETTh2}}} &
				\multicolumn{2}{c}{\rotatebox{0}{\scalebox{1}{Electricity}}} &
				\multicolumn{2}{c}{\rotatebox{0}{\scalebox{1}{Traffic}}} & \multicolumn{2}{c}{\rotatebox{0}{\scalebox{1}{Weather}}} &  \multicolumn{2}{c}{\rotatebox{0}{\scalebox{1}{ILI}}} \\
				\cmidrule(lr){0-0} \cmidrule(lr){2-3} \cmidrule(lr){4-5}\cmidrule(lr){6-7} \cmidrule(lr){8-9}\cmidrule(lr){10-11}\cmidrule(lr){12-13}\cmidrule(lr){14-15}\cmidrule(lr){16-17}
				\multicolumn{1}{c}{\scalebox{1}{Metric}} & \scalebox{1}{MSE} & \scalebox{1}{MAE} & \scalebox{1}{MSE} & \scalebox{1}{MAE} & \scalebox{1}{MSE} & \scalebox{1}{MAE} & \scalebox{1}{MSE} & \scalebox{1}{MAE} & \scalebox{1}{MSE} & \scalebox{1}{MAE} & \scalebox{1}{MSE} & \scalebox{1}{MAE} & \scalebox{1}{MSE} & \scalebox{1}{MAE} & \scalebox{1}{MSE} & \scalebox{1}{MAE} \\
				\toprule
				\scalebox{1}{High} &{\scalebox{1}{0.384}} &{\scalebox{1}{0.411}}  &\scalebox{1}{0.273} &\scalebox{1}{0.338} & \scalebox{1}{0.436} & \scalebox{1}{0.457} &\scalebox{1}{0.377} &\scalebox{1}{0.413} &\scalebox{1}{0.179} &\scalebox{1}{0.275} &\scalebox{1}{0.419} &\scalebox{1}{0.293} &\scalebox{1}{0.240} &\scalebox{1}{0.278} &\scalebox{1}{1.842} &\scalebox{1}{0.879}\\
				\midrule
				\scalebox{1}{Random} &\scalebox{1}{0.355} &\scalebox{1}{0.396} &\underline{\scalebox{1}{0.256}} &\scalebox{1}{0.321} & \scalebox{1}{0.418} & \scalebox{1}{0.440} &\underline{\scalebox{1}{0.337}} &\underline{\scalebox{1}{0.390}} &\scalebox{1}{0.169} &\scalebox{1}{0.262} &\underline{\scalebox{1}{0.394}} &\underline{0.269} &\underline{0.225} &\scalebox{1}{0.271} &\underline{1.458} &\underline{0.799}\\
				\midrule
				\scalebox{1}{Low} &\underline{\scalebox{1}{0.352}} &\underline{\scalebox{1}{0.390}} & 0.258 &\underline{\scalebox{1}{0.319}} & \underline{0.416} & \underline{0.437} &\scalebox{1}{0.340} &\scalebox{1}{0.392} &\underline{\scalebox{1}{0.165}} &\underline{\scalebox{1}{0.250}} & \scalebox{1}{0.397} & \scalebox{1}{0.273} & 0.227 &\underline{0.266} &\scalebox{1}{1.491} &{0.802}\\
				\midrule
				\scalebox{1}{Un-salient}&\textbf{\scalebox{1}{0.343}} &\textbf{\scalebox{1}{0.378}}  &\textbf{\scalebox{1}{0.246}} &\textbf{\scalebox{1}{0.313}} &\textbf{\scalebox{1}{0.413}} 
				&\textbf{\scalebox{1}{0.432}} &\textbf{\scalebox{1}{0.331}} &\textbf{\scalebox{1}{0.384}} &\textbf{\scalebox{1}{0.158}} &\textbf{\scalebox{1}{0.253}} &\textbf{\scalebox{1}{0.388}} &\textbf{\scalebox{1}{0.265}} &\textbf{\scalebox{1}{0.221}} &\textbf{\scalebox{1}{0.260}} &\textbf{\scalebox{1}{1.431}} &\textbf{\scalebox{1}{0.792}} \\
				\bottomrule
			\end{tabular}
	\end{threeparttable}
\end{table*}

\begin{table*}[htbp]
	\caption{Ablation study of the operation for salient part.}
	\label{tab:drop}
	\centering
	\begin{threeparttable}
		\renewcommand{\multirowsetup}{\centering}
		\setlength{\tabcolsep}{2pt}
		\begin{tabular}{c|cc|cc|cc|cc|cc|cc|cc|cc}
			\toprule
			\multicolumn{1}{c}{\scalebox{1}{Dataset}} & 
			\multicolumn{2}{c}{\rotatebox{0}{\scalebox{1}{ETTm1}}} &
			\multicolumn{2}{c}{\rotatebox{0}{\scalebox{1}{ETTm2}}} &
			\multicolumn{2}{c}{\rotatebox{0}{\scalebox{1}{ETTh1}}} &
			\multicolumn{2}{c}{\rotatebox{0}{\scalebox{1}{ETTh2}}} &
			\multicolumn{2}{c}{\rotatebox{0}{\scalebox{1}{Electricity}}} &
			\multicolumn{2}{c}{\rotatebox{0}{\scalebox{1}{Traffic}}} & \multicolumn{2}{c}{\rotatebox{0}{\scalebox{1}{Weather}}} & \multicolumn{2}{c}{\rotatebox{0}{\scalebox{1}{ILI}}} \\
			\cmidrule(lr){0-0} \cmidrule(lr){2-3} \cmidrule(lr){4-5}\cmidrule(lr){6-7} \cmidrule(lr){8-9}\cmidrule(lr){10-11}\cmidrule(lr){12-13}\cmidrule(lr){14-15}\cmidrule(lr){16-17}
			\multicolumn{1}{c}{\scalebox{1}{Metric}} & \scalebox{1}{MSE} & \scalebox{1}{MAE} & \scalebox{1}{MSE} & \scalebox{1}{MAE} & \scalebox{1}{MSE} & \scalebox{1}{MAE} & \scalebox{1}{MSE} & \scalebox{1}{MAE} & \scalebox{1}{MSE} & \scalebox{1}{MAE} & \scalebox{1}{MSE} & \scalebox{1}{MAE} & \scalebox{1}{MSE} & \scalebox{1}{MAE} & \scalebox{1}{MSE} & \scalebox{1}{MAE} \\
			\toprule
			\textit{drop} & 0.365 & 0.399 & 0.271 & 0.334 & 0.422 & 0.439 & 0.362 & 0.410 & 0.169 & 0.267 & 0.398 & 0.272 & 0.250 & 0.288 & 1.477 & 0.832 \\
			\midrule
			\textit{keep} &\textbf{\scalebox{1}{0.343}} &\textbf{\scalebox{1}{0.378}}  &\textbf{\scalebox{1}{0.246}} &\textbf{\scalebox{1}{0.313}} &\textbf{\scalebox{1}{0.413}} 
			&\textbf{\scalebox{1}{0.432}} &\textbf{\scalebox{1}{0.331}} &\textbf{\scalebox{1}{0.384}} &\textbf{\scalebox{1}{0.158}} &\textbf{\scalebox{1}{0.253}} &\textbf{\scalebox{1}{0.388}} &\textbf{\scalebox{1}{0.265}} &\textbf{\scalebox{1}{0.221}} &\textbf{\scalebox{1}{0.260}} &\textbf{\scalebox{1}{1.431}} &\textbf{\scalebox{1}{0.792}} \\
			\bottomrule
		\end{tabular}
	\end{threeparttable}
\end{table*}

\begin{table*}[htbp]
	\caption{Ablation study of the dynamic filter generator.}
	\label{tab:filter}
	\centering
	\begin{threeparttable}
		\renewcommand{\multirowsetup}{\centering}
		\setlength{\tabcolsep}{2pt}
		\begin{tabular}{c|cc|cc|cc|cc|cc|cc|cc|cc}
			\toprule
			\multicolumn{1}{c}{\scalebox{1}{Dataset}} & 
			\multicolumn{2}{c}{\rotatebox{0}{\scalebox{1}{ETTm1}}} &
			\multicolumn{2}{c}{\rotatebox{0}{\scalebox{1}{ETTm2}}} &
			\multicolumn{2}{c}{\rotatebox{0}{\scalebox{1}{ETTh1}}} &
			\multicolumn{2}{c}{\rotatebox{0}{\scalebox{1}{ETTh2}}} &
			\multicolumn{2}{c}{\rotatebox{0}{\scalebox{1}{Electricity}}} &
			\multicolumn{2}{c}{\rotatebox{0}{\scalebox{1}{Traffic}}} & \multicolumn{2}{c}{\rotatebox{0}{\scalebox{1}{Weather}}} & \multicolumn{2}{c}{\rotatebox{0}{\scalebox{1}{ILI}}} \\
			\cmidrule(lr){0-0} \cmidrule(lr){2-3} \cmidrule(lr){4-5}\cmidrule(lr){6-7} \cmidrule(lr){8-9}\cmidrule(lr){10-11}\cmidrule(lr){12-13}\cmidrule(lr){14-15}\cmidrule(lr){16-17}
			\multicolumn{1}{c}{\scalebox{1}{Metric}} & \scalebox{1}{MSE} & \scalebox{1}{MAE} & \scalebox{1}{MSE} & \scalebox{1}{MAE} & \scalebox{1}{MSE} & \scalebox{1}{MAE} & \scalebox{1}{MSE} & \scalebox{1}{MAE} & \scalebox{1}{MSE} & \scalebox{1}{MAE} & \scalebox{1}{MSE} & \scalebox{1}{MAE} & \scalebox{1}{MSE} & \scalebox{1}{MAE} & \scalebox{1}{MSE} & \scalebox{1}{MAE} \\
			\toprule
			\textit{deep} & 0.342 & 0.376 & 0.247 & 0.313 & 0.414 & 0.434 & 0.330 & 0.384 & 0.158 & 0.253 & 0.389 & 0.263 & 0.222 & 0.260 & 1.432 & 0.795 \\
			\midrule
			\textit{linear} &0.343 &0.378  &0.246 &0.313 &{0.413} 
			&{0.432} &{0.331} &{0.384} &{0.158} &{0.253} &{0.388} &{0.265} &{0.221} &{0.260} &{1.431} &{0.792} \\
			\bottomrule
		\end{tabular}
	\end{threeparttable}
\end{table*}

\section{More Ablation Studies}
\subsection{Ablation of Frequency Selection Strategies} 
\label{abla:selection_strategy}
To prove the effectiveness of our selection strategy, we compare four strategies: our un-salient frequency selection, random selection \cite{zhou2022fedformer} which randomly select $40\%$ lowest frequency and $10\%$ highes frequency follow \cite{Zhou2022FiLMFI}, low-frequency selection \cite{riad2022diffstride,Zhou2022FiLMFI}, and high-frequency selection strategies \cite{Zhou2022FiLMFI}.
As shown in Table \ref{tab:selecltion}, our salient frequency selection strategy outperforms other methods consistently which shows that although low frequency modes are important, the un-salinet modes in the high frequency part still contributes to the time series modeling. 

\subsection{Ablation Study of the operation for salient part}
We keep the salient part unchanged since even though these parts have low SNR, they still contribute to the feature in the time domain at each location. 
In this section, we verify this by comparing two operations: \textit{keep} and \textit{drop}, where \textit{keep} means we keep the non-salient parts in the frequency domain unchanged, and \textit{drop} means we set the non-salient parts within the frequency domain to $0$. As shown in Table \ref{tab:drop}, keeping the non-salient parts in the frequency domain still outperforms the \textit{drop} operation, and the \textit{drop} operation continuously decreases the performance; this is mainly because of the Picket Fence Effect of the FFT \cite{2008Eliminating}.

\subsection{Ablation of the dynamic filter generator}
We simply use a linear layer to generate the dynamic kernel which terms \textit{linear} since linear model can already extract trend and periodicity information for time series especially on the embed feature follow \cite{Zeng2022AreTE}. In this section, we explore to introduce more non-linearity to the kernel generator which terms \textit{deep}, where the structure of the sub-network is \textit{linear-gelu-linear}. However, the \textit{deep} variant shows limit performance improvement.

\begin{table*}[htbp]
	\caption{Effective of kernel parameter initialization.}
	\label{tab:init}
	\centering
	\begin{threeparttable}
		\renewcommand{\multirowsetup}{\centering}
		\setlength{\tabcolsep}{2pt}
		\begin{tabular}{c|cc|cc|cc|cc|cc|cc|cc|cc}
			\toprule
			\multicolumn{1}{c}{\scalebox{1}{Dataset}} & 
			\multicolumn{2}{c}{\rotatebox{0}{\scalebox{1}{ETTm1}}} &
			\multicolumn{2}{c}{\rotatebox{0}{\scalebox{1}{ETTm2}}} &
			\multicolumn{2}{c}{\rotatebox{0}{\scalebox{1}{ETTh1}}} &
			\multicolumn{2}{c}{\rotatebox{0}{\scalebox{1}{ETTh2}}} &
			\multicolumn{2}{c}{\rotatebox{0}{\scalebox{1}{Electricity}}} &
			\multicolumn{2}{c}{\rotatebox{0}{\scalebox{1}{Traffic}}} & \multicolumn{2}{c}{\rotatebox{0}{\scalebox{1}{Weather}}} & \multicolumn{2}{c}{\rotatebox{0}{\scalebox{1}{ILI}}} \\
			\cmidrule(lr){0-0} \cmidrule(lr){2-3} \cmidrule(lr){4-5}\cmidrule(lr){6-7} \cmidrule(lr){8-9}\cmidrule(lr){10-11}\cmidrule(lr){12-13}\cmidrule(lr){14-15}\cmidrule(lr){16-17}
			\multicolumn{1}{c}{\scalebox{1}{Metric}} & \scalebox{1}{MSE} & \scalebox{1}{MAE} & \scalebox{1}{MSE} & \scalebox{1}{MAE} & \scalebox{1}{MSE} & \scalebox{1}{MAE} & \scalebox{1}{MSE} & \scalebox{1}{MAE} & \scalebox{1}{MSE} & \scalebox{1}{MAE} & \scalebox{1}{MSE} & \scalebox{1}{MAE} & \scalebox{1}{MSE} & \scalebox{1}{MAE} & \scalebox{1}{MSE} & \scalebox{1}{MAE} \\
			\toprule
			\textit{random} & {\scalebox{1}{0.372}} &{\scalebox{1}{0.403}}  &\scalebox{1}{0.274} &\scalebox{1}{0.336} & {\scalebox{1}{0.431}} & \scalebox{1}{0.453} &\scalebox{1}{0.382} &{\scalebox{1}{0.430}} &\scalebox{1}{0.188} &\scalebox{1}{0.277} &{\scalebox{1}{0.428}} &{\scalebox{1}{0.299}} &\scalebox{1}{0.235} &{\scalebox{1}{0.270}} &\scalebox{1}{1.927} &{\scalebox{1}{0.913}}\\
			\midrule
			\textit{one} & 0.372 & 0.403 & 0.275 &  0.336 &  0.432 &  0.452 &  0.383 &  0.430 &  0.186 &  0.276 &  0.429 &  0.299 &  0.233 &  0.270 &  1.924 &  0.915 \\
			\midrule
			\textit{sgconv} & 0.371 &  0.402 &  0.278 &  0.337 &  0.431 &  0.452 &  0.382 &  0.430 &  0.187 &  0.277 &  0.428 &  0.298 &  0.235 &  0.270 &  1.923 &  0.911  \\
			\bottomrule
		\end{tabular}
	\end{threeparttable}
\end{table*}

\begin{table*}[htbp]
	\centering
	\caption{The results of noise injection experiment. A $0.1\times\mathcal{N}(0,1)$ Gaussian noise is introduced. We conduct four sets of experiments with/without noise in training and test phases. The metric of variants is presented in relative value, where $\uparrow$ indicates degraded performance, and $\downarrow$ indicates improved performance.}
	\label{tab:noise}
	\vskip 0.05in
	\setlength{\tabcolsep}{3pt}
	\begin{tabular}{c|c|cc|cc}
		\toprule
		\multicolumn{2}{c|}{Training} &
		\multicolumn{2}{c|}{w.o. noise} & \multicolumn{2}{c}{noise}  \\
		\midrule
		\multicolumn{2}{c|}{Testing} & 
		w.o. noise & noise & w.o. noise & noise \\
		\toprule
		\multirow{4}{*}{\makecell[c]{Long-term\\Forecasting\\(MSE)}} & ETTm1
		& 0.343 & $\uparrow$0.29\% & $\downarrow$0.57\% & $\uparrow$0.62\% \\ 
		& ETTm2 & 0.246 & $\downarrow$0.60\% & $\downarrow$0.91\% & $\downarrow$0.09\% \\
		& ETTh1 & 0.413 & $\downarrow$0.48\% & $\uparrow$0.11\% & $\uparrow$0.82\% \\
		& ETTh2 & 0.331 & $\uparrow$0.24\% & $\uparrow$0.08\% & $\downarrow$0.16\% \\ 
		\midrule
		\multirow{2}{*}{\makecell[c]{Classification\\(ACC)}} & EthanolConcentration
		& 36.6 & $\downarrow$0.27$\%$ & $\downarrow$0.82\% & $\uparrow$0.27\% \\ 
		& FaceDetection & 69.5 & $\downarrow$0.14\% & $\uparrow$0.36\% & 0.00\% \\
		\midrule
		\multirow{2}{*}{\makecell[c]{Anomaly Detection\\(F1 Score)}} & SMAP & 70.44 & $\downarrow$0.14\% & $\uparrow$0.21\% & $\downarrow$0.04\% \\ 
		& MSL & 84.43 & 0.00\% & 0.00\% & 0.00\% \\
		\bottomrule
	\end{tabular}
\end{table*}

\subsection{Effective of kernel parameter initialization}
\label{appen:init}
In Section \ref{sec:analysis}, SGConv is compared with our method, the results show that their proposed principles have minimal improvement for time series task which are conflict to the results on the LRA benchmark. To further understand the reason behind the phenomenon in depth, we compare different parameter initialization methods for the global kernel: \textit{random} initialization from normal Gaussian distribution, unified initialization which initialize all parameters to value \textit{one}, and \textit{sgconv} initialization method. The similar results in Table \ref{tab:init} show that the baseline model on time series task is robust to kernel initialization. 

\section{Noise Injection Experiment}
To verify our model's robustness, we conduct noise injection experiments by adding Gaussian noise with $0.1\times\mathcal{N}(0,1)$ in the training/test stage. The results in Table \ref{tab:noise} show that adding noise has a limited effect on our model’s performance, the deterioration is less than 1\% in the worst case. Thanks to the introducing of the sparse operation on the un-salient part in the frequency domain, the model’s robustness is consistent across various tasks. Moreover, as shown in the last line, adding noise even on the train or test stage has no effect on the MSL dataset.

\section{Parameter Sensitivity Analysis}
\subsection{Hyper-parameter sensitivity analysis}
\begin{figure*}[h]{}
	\caption{Analysis of hyper-parameter sensitivity.}
	\label{fig_hyper}
	\centering
	\subfigure[number of backbone layers]{
		\label{hyper_1}
		\includegraphics[width=0.3\linewidth]{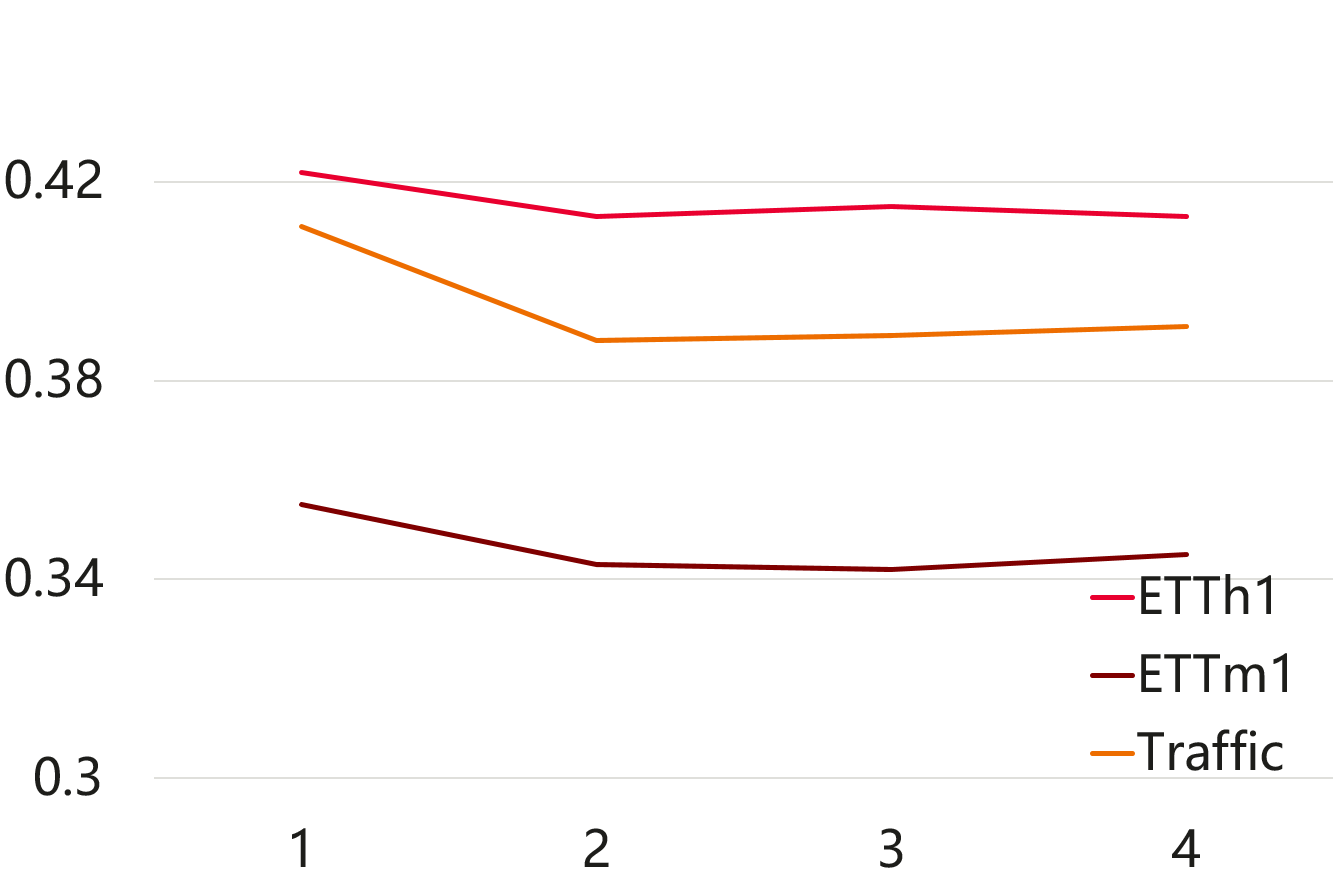}
	}
	\subfigure[number of mlp ratio]{
		\label{hyper_2}
		\includegraphics[width=0.3\linewidth]{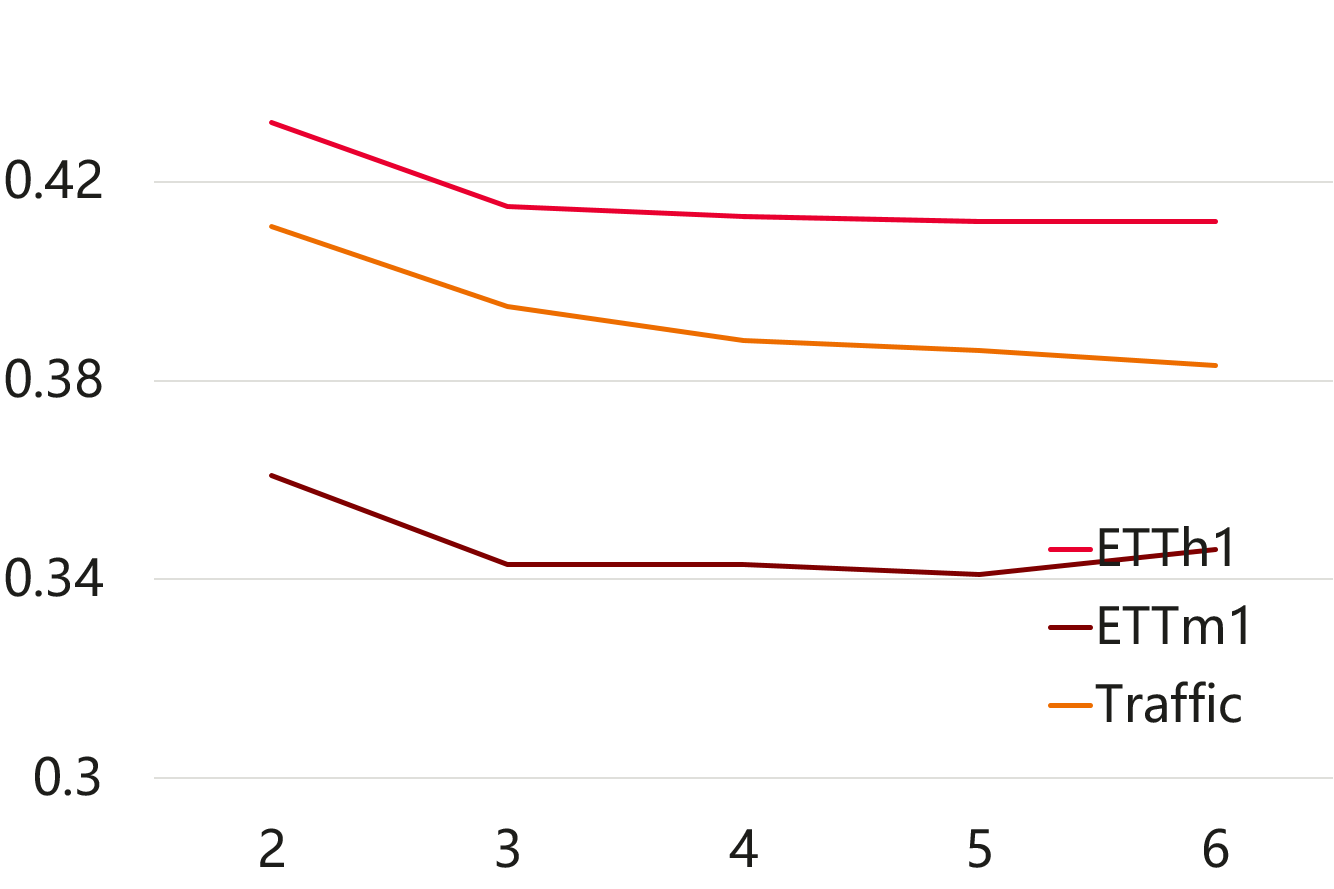}
	}
	\subfigure[input length]{
		\label{hyper_3}
		\includegraphics[width=0.3\linewidth]{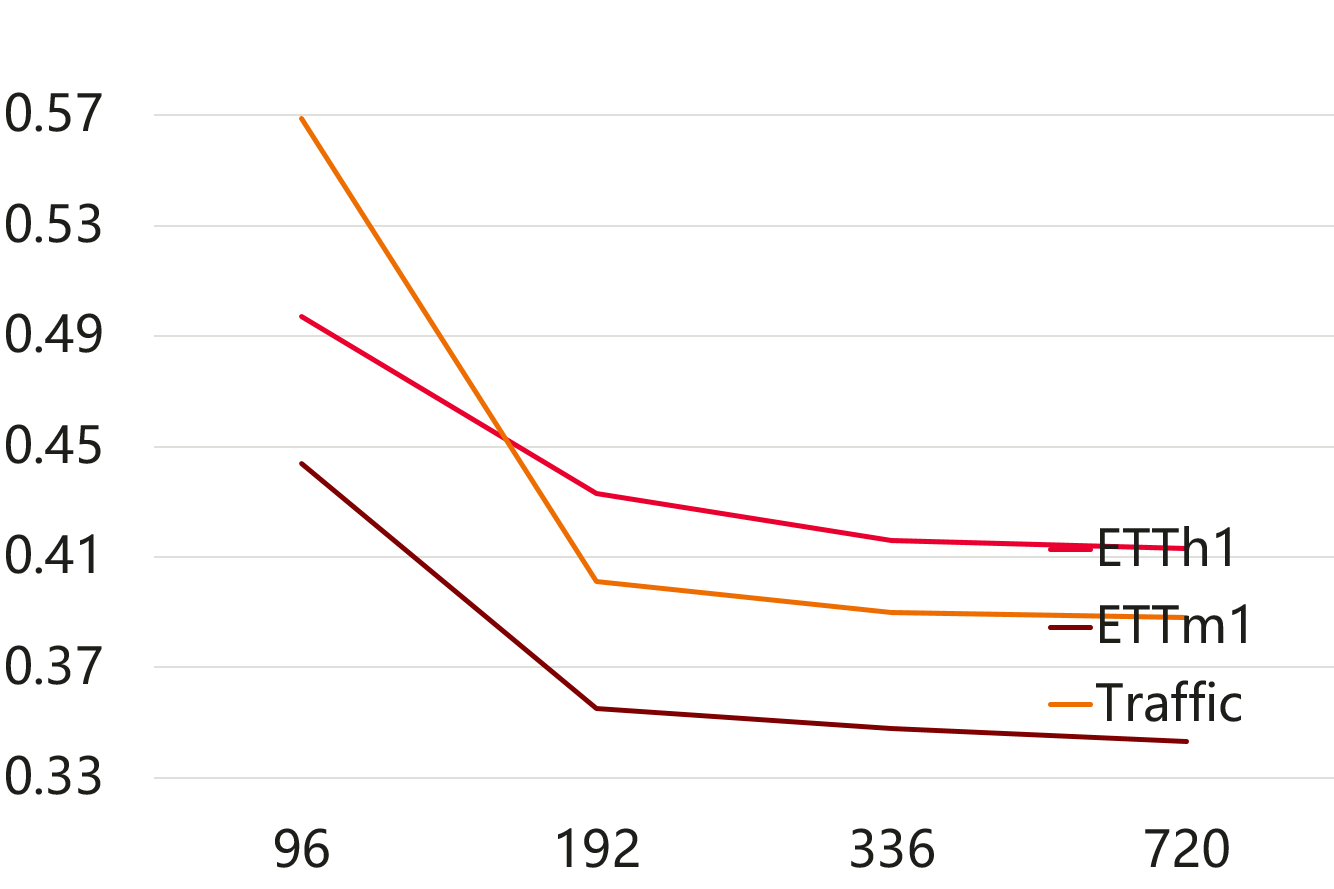}
	}
\end{figure*}

We conduct a hyperparameter sensitivity analysis focusing on the important hyper-parameters within ODFL: namely, the number of backbone model layers, the mlp ratio, the time series input length $T$. The correlated results can be found in Figure \ref{fig_hyper}. We can find that our proposed ODFL can present a stable performance under different choices of hyper-parameters. And it shows that the performance first increases and then slightly decreases with respect to the increase of the
model size because a large size improves the fitting ability of our ODFL but may
easily lead to overfitting especially when the model is too large. Besides, the model can capture more temporal information with longer input sequence.

\subsection{Experiment error bar under different seeds}
All experiments have been conducted three times with different seeds, and we present the standard deviations of our model. In Table \ref{tab:std}, the average MSE and MAE have been reported across four ETT datasets, complete with standard deviations. The result shows that random seeds make very little impact on the ODFL.

\begin{table*}[htbp]
	\caption{Standard deviations of our ODFL.}
	\label{tab:std}
	\centering
	\begin{threeparttable}
		\renewcommand{\multirowsetup}{\centering}
		\setlength{\tabcolsep}{1.2pt}
		\begin{tabular}{c|cc|cc|cc|cc}
			\toprule
			\multicolumn{1}{c}{\scalebox{1}{Dataset}} & 
			\multicolumn{2}{c}{\rotatebox{0}{\scalebox{1}{ETTm1}}} &
			\multicolumn{2}{c}{\rotatebox{0}{\scalebox{1}{ETTm2}}} &
			\multicolumn{2}{c}{\rotatebox{0}{\scalebox{1}{ETTh1}}} &
			\multicolumn{2}{c}{\rotatebox{0}{\scalebox{1}{ETTh2}}} \\
			\cmidrule(lr){0-0} \cmidrule(lr){2-3} \cmidrule(lr){4-5}\cmidrule(lr){6-7} \cmidrule(lr){8-9}
			\multicolumn{1}{c}{\scalebox{1}{Metric}} & \scalebox{1}{MSE} & \scalebox{1}{MAE} & \scalebox{1}{MSE} & \scalebox{1}{MAE} & \scalebox{1}{MSE} & \scalebox{1}{MAE} & \scalebox{1}{MSE} & \scalebox{1}{MAE} \\
			\toprule
			Result &0.343$\pm$0.005 &0.378$\pm$0.003 &0.246$\pm$0.001 &0.313$\pm$0.004 &0.413$\pm$0.006 
			&{0.432$\pm$0.008} &{0.331$\pm$0.003} &{0.384$\pm$0.001} \\
			\bottomrule
		\end{tabular}
	\end{threeparttable}
\end{table*}

\section{Visualization}
\label{appen:vis_kernel}
\begin{figure*}[]
	\begin{center}
		\includegraphics[width=0.99\linewidth]{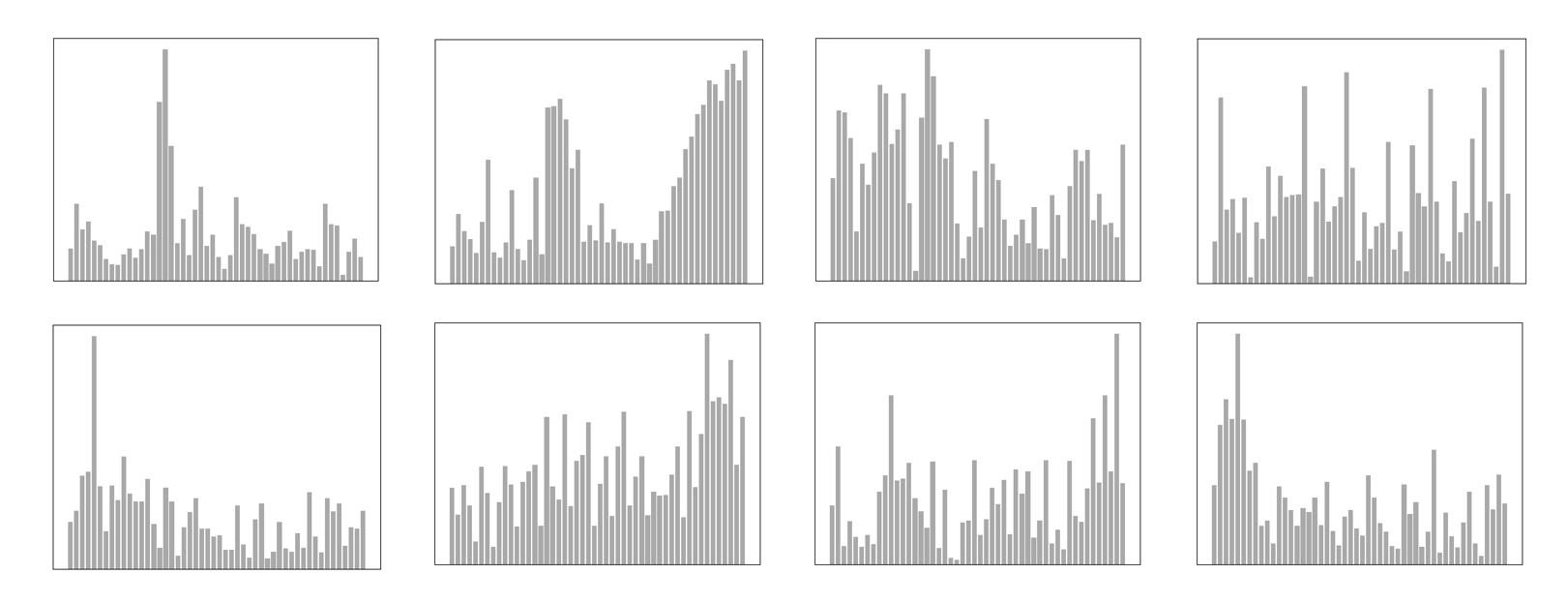}
		\caption{Visualization of our learned kernel. The horizontal axis represents index, the vertical axis represents the amplitude value of the kernel $\bm{K}$.}
		\label{fig_kernel}
	\end{center}
\end{figure*}

We provide the visualization of our learned kernel in the frequency domain on ETTh1 dataset in Figure \ref{fig_kernel}. The visualization shows that our kernel learns diversified and preferential frequency bands. Also, some of our kernels has learned the decaying structure introduced by SGConv \cite{Li2022WhatMC}.

\section{Full results}
\label{appendix:full}
Due to the space limitation of the main text, we place the full results of all experiments in the following: long-term forecasting in Table~\ref{tab:full_forecasting_results}, imputation in Table~\ref{tab:full_imputation_results},  short-term forecasting in Table~\ref{tab:full_forecasting_results_m4}, anomaly detection in Table~\ref{tab:full_anomaly_results}, and classification in Table~\ref{tab:full_classification_results}.

\begin{table*}[htbp]
	\caption{Full results for the long-term forecasting task.}
	\label{tab:full_forecasting_results}
	\centering
	\begin{threeparttable}
			\renewcommand{\multirowsetup}{\centering}
			\setlength{\tabcolsep}{1pt}
\begin{tabular}{c|c|cc|cc|cc|cc|cc|cc|cc|cc|cc|cc|cc|cc}
				\toprule
				\multicolumn{2}{c}{\multirow{2}{*}{\scalebox{0.7}{Models}}} & 
				\multicolumn{2}{c}{\rotatebox{0}{\scalebox{0.7}{ODFL}}} &
				\multicolumn{2}{c}{\rotatebox{0}{\scalebox{0.7}{TimesNet}}} &
				\multicolumn{2}{c}{\rotatebox{0}{\scalebox{0.7}{ETS.}}} &
				\multicolumn{2}{c}{\rotatebox{0}{\scalebox{0.7}{LightTS}}} &
				\multicolumn{2}{c}{\rotatebox{0}{\scalebox{0.7}{DLinear}}} &
				\multicolumn{2}{c}{\rotatebox{0}{\scalebox{0.7}{FED.}}} & \multicolumn{2}{c}{\rotatebox{0}{\scalebox{0.7}{PatchTST}}} & \multicolumn{2}{c}{\rotatebox{0}{\scalebox{0.7}{OFA}}} & \multicolumn{2}{c}{\rotatebox{0}{\scalebox{0.7}{Cross.}}} &  \multicolumn{2}{c}{\rotatebox{0}{\scalebox{0.7}{SCINet}}} & \multicolumn{2}{c}{\rotatebox{0}{\scalebox{0.7}{MICN}}}  & \multicolumn{2}{c}{\rotatebox{0}{\scalebox{0.7}{FreTS}}}  \\
				\multicolumn{2}{c}{} &
				\multicolumn{2}{c}{\scalebox{0.7}{(Ours)}} & \multicolumn{2}{c}{\scalebox{0.7}{(2023)}} & 
				\multicolumn{2}{c}{\scalebox{0.7}{(2022)}} &
				\multicolumn{2}{c}{\scalebox{0.7}{(2022)}} &
				\multicolumn{2}{c}{\scalebox{0.7}{(2023)}} & \multicolumn{2}{c}{\scalebox{0.7}{(2022)}} & \multicolumn{2}{c}{\scalebox{0.7}{(2023)}} & \multicolumn{2}{c}{\scalebox{0.7}{(2023)}} & \multicolumn{2}{c}{\scalebox{0.7}{(2023)}} &  \multicolumn{2}{c}{\scalebox{0.7}{(2022)}} & \multicolumn{2}{c}{\scalebox{0.7}{(2023)}} & \multicolumn{2}{c}{\scalebox{0.7}{(2023)}}  \\
\bottomrule
			\end{tabular}
	\end{threeparttable}
\end{table*}

\begin{table*}[htbp]
	\caption{Full results for the short-term forecasting task.}
	\label{tab:full_forecasting_results_m4}
	\centering
	\begin{threeparttable}
			\renewcommand{\multirowsetup}{\centering}
			\setlength{\tabcolsep}{1pt}
			\begin{tabular}{cc|c|ccccccccccccccccccc}
				\toprule
				\multicolumn{3}{c}{\multirow{2}{*}{Models}} & 
				\multicolumn{1}{c}{\rotatebox{0}{\scalebox{0.7}{ODFL}}} &
				\multicolumn{1}{c}{\rotatebox{0}{\scalebox{0.7}{TimesNet}}} &
				\multicolumn{1}{c}{\rotatebox{0}{\scalebox{0.7}{{N-HiTS}}}} &
				\multicolumn{1}{c}{\rotatebox{0}{\scalebox{0.7}{{N-BEATS}}}} &
				\multicolumn{1}{c}{\rotatebox{0}{\scalebox{0.7}{ETS.}}} &
				\multicolumn{1}{c}{\rotatebox{0}{\scalebox{0.7}{LightTS}}} &
				\multicolumn{1}{c}{\rotatebox{0}{\scalebox{0.7}{DLinear}}} &
				\multicolumn{1}{c}{\rotatebox{0}{\scalebox{0.7}{FED.}}} & \multicolumn{1}{c}{\rotatebox{0}{\scalebox{0.7}{Stationary}}} & \multicolumn{1}{c}{\rotatebox{0}{\scalebox{0.7}{Cross.}}} & \multicolumn{1}{c}{\rotatebox{0}{\scalebox{0.7}{PatchTST}}} &  \multicolumn{1}{c}{\rotatebox{0}{\scalebox{0.7}{MICN}}} & \multicolumn{1}{c}{\rotatebox{0}{\scalebox{0.7}{SCINet}}} & \multicolumn{1}{c}{\rotatebox{0}{\scalebox{0.7}{OFA}}} & \multicolumn{1}{c}{\rotatebox{0}{\scalebox{0.7}{FreTS}}}
				\\
				\multicolumn{1}{c}{}&\multicolumn{1}{c}{} &\multicolumn{1}{c}{} &
				\multicolumn{1}{c}{\scalebox{0.7}{(Ours)}} &
				\multicolumn{1}{c}{\scalebox{0.7}{(2023)}} &
				\multicolumn{1}{c}{\scalebox{0.7}{(2022)}} &
				\multicolumn{1}{c}{\scalebox{0.7}{(2019)}} &
				\multicolumn{1}{c}{\scalebox{0.7}{(2022)}} &
				\multicolumn{1}{c}{\scalebox{0.7}{(2022)}} &
				\multicolumn{1}{c}{\scalebox{0.7}{(2023)}} & \multicolumn{1}{c}{\scalebox{0.7}{(2022)}} & \multicolumn{1}{c}{\scalebox{0.7}{(2022)}} & \multicolumn{1}{c}{\scalebox{0.7}{(2023)}} & \multicolumn{1}{c}{\scalebox{0.7}{(2023)}} &  \multicolumn{1}{c}{\scalebox{0.7}{(2023)}} & \multicolumn{1}{c}{\scalebox{0.7}{(2023)}} & \multicolumn{1}{c}{\scalebox{0.7}{(2023)}} & \multicolumn{1}{c}{\scalebox{0.7}{(2023)}}\\
\bottomrule
			\end{tabular}
	\end{threeparttable}
\end{table*}

\begin{table*}[htbp]
	\caption{Full results for the imputation task.}\label{tab:full_imputation_results}
	\centering
		\begin{threeparttable}
				\renewcommand{\multirowsetup}{\centering}
				\setlength{\tabcolsep}{1pt}
				\begin{tabular}{c|c|cc|cc|cc|cc|cc|cc|cc|cc|cc|cc|cc|cc}
					\toprule
					\multicolumn{2}{c}{\multirow{2}{*}{Models}} & 
					\multicolumn{2}{c}{\rotatebox{0}{\scalebox{0.7}{ODFL}}} &
					\multicolumn{2}{c}{\rotatebox{0}{\scalebox{0.7}{TimesNet}}} &
					\multicolumn{2}{c}{\rotatebox{0}{\scalebox{0.7}{ETS.}}} &
					\multicolumn{2}{c}{\rotatebox{0}{\scalebox{0.7}{LightTS}}} &
					\multicolumn{2}{c}{\rotatebox{0}{\scalebox{0.7}{DLinear}}} &
					\multicolumn{2}{c}{\rotatebox{0}{\scalebox{0.7}{FED.}}} & \multicolumn{2}{c}{\rotatebox{0}{\scalebox{0.7}{PatchTST}}} & \multicolumn{2}{c}{\rotatebox{0}{\scalebox{0.7}{Cross.}}} & \multicolumn{2}{c}{\rotatebox{0}{\scalebox{0.7}{MICN}}} &  \multicolumn{2}{c}{\rotatebox{0}{\scalebox{0.7}{SCINet}}} & \multicolumn{2}{c}{\rotatebox{0}{\scalebox{0.7}{OFA}}} & \multicolumn{2}{c}{\rotatebox{0}{\scalebox{0.7}{FreTS}}}
					\\
					\multicolumn{2}{c}{} & 
					\multicolumn{2}{c}{\scalebox{0.7}{(Ours)}}& 
					\multicolumn{2}{c}{\scalebox{0.7}{(2023)}} & 
					\multicolumn{2}{c}{\scalebox{0.7}{(2022)}} &
					\multicolumn{2}{c}{\scalebox{0.7}{(2022)}} &
					\multicolumn{2}{c}{\scalebox{0.7}{(2023)}} & \multicolumn{2}{c}{\scalebox{0.7}{(2022)}} & \multicolumn{2}{c}{\scalebox{0.7}{(2023)}} & \multicolumn{2}{c}{\scalebox{0.7}{(2023)}} & \multicolumn{2}{c}{\scalebox{0.7}{(2023)}} &  \multicolumn{2}{c}{\scalebox{0.7}{(2022)}} & \multicolumn{2}{c}{\scalebox{0.7}{(2023)}} 
					& \multicolumn{2}{c}{\scalebox{0.7}{(2023)}} 
					\\
\bottomrule
				\end{tabular}
	\end{threeparttable}
\end{table*}

\begin{table*}[tbp]
	\caption{Full results for the anomaly detection task. The P, R and F1 represent the precision, recall and F1-score (\%) respectively. F1-score is the harmonic mean of precision and recall. A higher value of P, R and F1 indicates a better performance.}\label{tab:full_anomaly_results}
	\vskip 0.05in
	\centering
	\begin{threeparttable}
			\renewcommand{\multirowsetup}{\centering}
			\setlength{\tabcolsep}{3pt}
			\begin{tabular}{lc|ccc|ccc|ccc|ccc|ccc|c}
				\toprule
				\multicolumn{2}{c}{\scalebox{0.7}{Datasets}} & 
				\multicolumn{3}{c}{\scalebox{0.7}{\rotatebox{0}{SMD}}} &
				\multicolumn{3}{c}{\scalebox{0.7}{\rotatebox{0}{MSL}}} &
				\multicolumn{3}{c}{\scalebox{0.7}{\rotatebox{0}{SMAP}}} &
				\multicolumn{3}{c}{\scalebox{0.7}{\rotatebox{0}{SWaT}}} & 
				\multicolumn{3}{c}{\scalebox{0.7}{\rotatebox{0}{PSM}}} & \scalebox{0.7}{Avg F1} \\
				\cmidrule(lr){0-1} \cmidrule(lr){3-5} \cmidrule(lr){6-8}\cmidrule(lr){9-11} \cmidrule(lr){12-14} \cmidrule(lr){15-17} \cmidrule(lr){18-18}
				\multicolumn{2}{c}{\scalebox{0.7}{Metrics}} & \scalebox{0.7}{P} & \scalebox{0.7}{R} & \scalebox{0.7}{F1} & \scalebox{0.7}{P} & \scalebox{0.7}{R} & \scalebox{0.7}{F1} & \scalebox{0.7}{P} & \scalebox{0.7}{R} & \scalebox{0.7}{F1} & \scalebox{0.7}{P} & \scalebox{0.7}{R} & \scalebox{0.7}{F1} & \scalebox{0.7}{P} & \scalebox{0.7}{R} & \scalebox{0.7}{F1} & \scalebox{0.7}{(\%)}\\
				\toprule
				\scalebox{0.7}{Anomaly} & \scalebox{0.7}{(2021)} %
				& \scalebox{0.7}{88.91} & \scalebox{0.7}{82.23} & \scalebox{0.7}{85.49}
				& \scalebox{0.7}{79.61} & \scalebox{0.7}{87.37} & \scalebox{0.7}{83.31} 
				& \scalebox{0.7}{91.85} & \scalebox{0.7}{58.11} & \scalebox{0.7}{{71.18}}
				& \scalebox{0.7}{72.51} & \scalebox{0.7}{97.32} & \scalebox{0.7}{83.10} 
				& \scalebox{0.7}{68.35} & \scalebox{0.7}{94.72} & \scalebox{0.7}{79.40} 
				& \scalebox{0.7}{80.50} \\
				\scalebox{0.7}{OFA} & \scalebox{0.7}{(2023)}
				& \scalebox{0.7}{88.89} & \scalebox{0.7}{84.98} & \underline{\scalebox{0.7}{86.89}}
				& \scalebox{0.7}{82.00} & \scalebox{0.7}{82.91} & \scalebox{0.7}{82.45}
				& \scalebox{0.7}{90.60} & \scalebox{0.7}{60.95} & \underline{\scalebox{0.7}{72.88}} 
				& \scalebox{0.7}{92.20} & \scalebox{0.7}{96.34} & \textbf{\scalebox{0.7}{94.23}} 
				& \scalebox{0.7}{98.62} & \scalebox{0.7}{95.68} & {\scalebox{0.7}{97.13}}
				&\textbf{\scalebox{0.7}{86.72}} \\
\bottomrule
			\end{tabular}
	\end{threeparttable}
\end{table*}

\begin{table*}[tbp]
	\caption{Full results for the classification task. We report the classification accuracy (\%) results.}
	\label{tab:full_classification_results}
	\centering
	\begin{threeparttable}
			\renewcommand{\multirowsetup}{\centering}
			\setlength{\tabcolsep}{1pt}
			\begin{tabular}{c|ccccccccccccccccccccc}
				\toprule
				\multirow{3}{*}{\scalebox{0.7}{Datasets}} 
				&\multicolumn{4}{c}{\scalebox{0.7}{CNN}} 
				&\multicolumn{2}{c}{\scalebox{0.7}{RNN}}
				&\multicolumn{3}{c}{\scalebox{0.7}{MLP}} 
				&\multicolumn{5}{c}{\scalebox{0.7}{Transformers}} 
				&\multicolumn{1}{c}{\scalebox{0.7}{Ours}}
				\\
				\cmidrule(lr){2-5}\cmidrule(lr){6-7}\cmidrule(lr){8-10}\cmidrule(lr){11-15}\cmidrule(lr){16-16}
				& \scalebox{0.7}{MICN} & \scalebox{0.7}{Rocket}  & \scalebox{0.7}{SCINet} & \scalebox{0.7}{TimesNet} &\scalebox{0.7}{LSTNet}  & \scalebox{0.7}{S4} & \scalebox{0.7}{DLinear} & \scalebox{0.7}{LightTS} &
				\scalebox{0.7}{FreTS} & \scalebox{0.7}{ETS.} & \scalebox{0.7}{PatchTST} & \scalebox{0.7}{Cross.} &  \scalebox{0.7}{FED.} & \scalebox{0.7}{OFA} & \scalebox{0.7}{ODFL} 
				\\
				& \scalebox{0.7}{(2023)} &  \scalebox{0.7}{(2020)} & 
				\scalebox{0.7}{(2022)} & 
				\scalebox{0.7}{(2023)} & 
				\scalebox{0.7}{(2018)} & 
				\scalebox{0.7}{(2022)} & 
				\scalebox{0.7}{(2023)} & 
				\scalebox{0.7}{(2022)} & 
				\scalebox{0.7}{(2023)} &
				\scalebox{0.7}{(2022)} &
				\scalebox{0.7}{(2023)} & 
				\scalebox{0.7}{(2023)} &
				\scalebox{0.7}{(2022)} & 
				\scalebox{0.7}{(2023)} & 
				\scalebox{0.7}{(Ours)} \\
				\toprule
				\scalebox{0.7}{EthanolConcentration} & \scalebox{0.7}{35.3} & \scalebox{0.7}{45.2}  & \scalebox{0.7}{34.4} & \scalebox{0.7}{35.7} & \scalebox{0.7}{39.9} & \scalebox{0.7}{31.1} &\scalebox{0.7}{36.2} &\scalebox{0.7}{29.7}  
				&\scalebox{0.7}{28.9} &\scalebox{0.7}{28.1} &\scalebox{0.7}{32.8} &\scalebox{0.7}{38.0} &\scalebox{0.7}{31.2} & \scalebox{0.7}{34.2} & \scalebox{0.7}{36.6} \\
\bottomrule
			\end{tabular}
	\end{threeparttable}
\end{table*}

\end{document}